\pdfoutput=1

\documentclass{article} 
\usepackage{iclr2027_conference,times}
\usepackage{booktabs}

\usepackage{amsmath,amsfonts,bm}

\def\eqref#1{equation~\ref{#1}}

\def\1{\bm{1}}

\DeclareMathAlphabet{\mathsfit}{\encodingdefault}{\sfdefault}{m}{sl}
\SetMathAlphabet{\mathsfit}{bold}{\encodingdefault}{\sfdefault}{bx}{n}

\def\gN{{\mathcal{N}}}

\newcommand{\E}{\mathbb{E}}

\newcommand{\R}{\mathbb{R}}

\newcommand{\bootReps}{2,000}
\newcommand{\critAucGap}{0.01}
\newcommand{\critCoefRMin}{0.99}
\newcommand{\critCoverageLevels}{50, 80 and 90\,\%}
\newcommand{\critCoverageTolPp}{3}
\newcommand{\critEceGap}{0.01}
\newcommand{\critHeadroomFlagNats}{0.02}
\newcommand{\critHeadroomMin}{0.95}
\newcommand{\critResPooled}{0.90}
\newcommand{\critResSubset}{0.85}
\newcommand{\exactMaxP}{5}
\newcommand{\exactNModelsMaxP}{1,450}

\newcommand{\finalCoefRProbeExactPanel}{0.9958}

\newcommand{\fitCpuOverGpu}{222}
\newcommand{\fitGpuName}{NVIDIA L40}
\newcommand{\fitMcmcOverGpu}{1,924}

\newcommand{\fitSecGpu}{0.03}
\newcommand{\fitSecGpuLoad}{0.92}
\newcommand{\fitSecGpuMax}{0.42}

\newcommand{\fitSecGpuN}{10,000}

\newcommand{\fitSecGpuPNinety}{0.20}

\newcommand{\fitSecGpuQOne}{0.01}
\newcommand{\fitSecGpuQThree}{0.10}
\newcommand{\gateAucGap}{\ensuremath{-0.0026}}

\newcommand{\gateAucGapArmVTwoMixFiveZeroNTwoKStrongNonllS}{\ensuremath{-0.0055}}

\newcommand{\gateCoefRExactPanel}{0.9931}

\newcommand{\gateEceGap}{0.0034}

\newcommand{\gateEceGapArmVTwoMixFiveZeroNTwoKStrongNonllS}{0.0005}

\newcommand{\gateResRatio}{0.9640}

\newcommand{\gateResRatioArmVTwoMixFiveZeroNTwoKStrongNonllS}{0.9050}

\newcommand{\hpAZero}{3}
\newcommand{\hpBZero}{1}
\newcommand{\hpBinaryRateHi}{0.5}
\newcommand{\hpBinaryRateLo}{0.2}
\newcommand{\hpCMax}{1.5}
\newcommand{\hpCMedian}{0.15}
\newcommand{\hpCMin}{0.01}
\newcommand{\hpCountShare}{0.5}
\newcommand{\hpCrossBlockWMax}{0.35}
\newcommand{\hpCrossLoadShare}{0.15}
\newcommand{\hpFBetaA}{1.5}
\newcommand{\hpFBetaB}{3}
\newcommand{\hpFZeroProb}{0.15}
\newcommand{\hpFactorFracInv}{3}
\newcommand{\hpHTwoCap}{0.98}
\newcommand{\hpHTwoConcHi}{40}
\newcommand{\hpHTwoConcLo}{4}
\newcommand{\hpHTwoHighCap}{0.995}
\newcommand{\hpHTwoHighMeanLo}{0.9}
\newcommand{\hpHTwoHighProb}{0.15}
\newcommand{\hpHTwoMeanHi}{0.92}
\newcommand{\hpHTwoMeanLo}{0.1}
\newcommand{\hpKIntMean}{1.5}
\newcommand{\hpKIntZero}{0.5}
\newcommand{\hpKMainMean}{3.5}
\newcommand{\hpKMainZero}{0.09}
\newcommand{\hpLikertLevelsMax}{7}
\newcommand{\hpLikertLevelsMin}{2}
\newcommand{\hpMargBinary}{0.1}
\newcommand{\hpMargCount}{0.05}
\newcommand{\hpMargGaussian}{0.6}
\newcommand{\hpMargLikert}{0.2}
\newcommand{\hpMargSkewed}{0.05}

\newcommand{\hpNegLoadProb}{0.3}
\newcommand{\hpPiIntHi}{0.4}
\newcommand{\hpPiIntLo}{0.05}
\newcommand{\hpPiMainHi}{0.6}
\newcommand{\hpPiMainLo}{0.02}
\newcommand{\hpRho}{0.5}
\newcommand{\hpSC}{0.6}
\newcommand{\hpTauZeroSq}{0.25}
\newcommand{\ipAcceptMaxPctRealX}{3.6\%}
\newcommand{\ipAcceptMinPctRealX}{0.013\%}

\newcommand{\ipAucMeanGlinternetRealX}{0.79}
\newcommand{\ipAucMeanHierNetRealX}{0.92}
\newcommand{\ipAucMeanLassoRealX}{0.82}

\newcommand{\ipAucMeanOursRealX}{0.96}
\newcommand{\ipAucMeanStabilityRefitRealX}{0.82}
\newcommand{\ipAucMeanSusieRealX}{0.85}

\newcommand{\ipAucNOursRealX}{9,974}

\newcommand{\ipAucQOneGlinternetRealX}{0.68}
\newcommand{\ipAucQOneHierNetRealX}{0.88}
\newcommand{\ipAucQOneLassoRealX}{0.72}

\newcommand{\ipAucQOneOursRealX}{0.95}
\newcommand{\ipAucQOneStabilityRefitRealX}{0.72}
\newcommand{\ipAucQOneSusieRealX}{0.76}

\newcommand{\ipAucQThreeGlinternetRealX}{0.97}
\newcommand{\ipAucQThreeHierNetRealX}{1.00}
\newcommand{\ipAucQThreeLassoRealX}{0.96}

\newcommand{\ipAucQThreeOursRealX}{1.00}
\newcommand{\ipAucQThreeStabilityRefitRealX}{0.95}
\newcommand{\ipAucQThreeSusieRealX}{1.00}

\newcommand{\ipBinWidthRealX}{0.1}
\newcommand{\ipCellsRealX}{25}

\newcommand{\ipDatasetsExactRealX}{2,299}

\newcommand{\ipDatasetsOursRealX}{10,000}

\newcommand{\ipDesignsPerCellMaxRealX}{220}
\newcommand{\ipDesignsPerCellMinRealX}{101}
\newcommand{\ipDiffAucExactRealX}{\ensuremath{-0.009}}
\newcommand{\ipDiffAucGlinternetRealX}{+0.173}
\newcommand{\ipDiffAucHierNetRealX}{+0.044}
\newcommand{\ipDiffAucLassoRealX}{+0.144}

\newcommand{\ipDiffAucLinearpfnStrongNonllRealX}{+0.005}
\newcommand{\ipDiffAucMcThreeRealX}{\ensuremath{-0.008}}
\newcommand{\ipDiffAucStabilityRefitRealX}{+0.147}
\newcommand{\ipDiffAucSusieRealX}{+0.110}
\newcommand{\ipDiffFOneExactRealX}{\ensuremath{-0.030}}
\newcommand{\ipDiffFOneGlinternetRealX}{+0.205}
\newcommand{\ipDiffFOneHierNetRealX}{+0.244}
\newcommand{\ipDiffFOneLassoRealX}{+0.293}

\newcommand{\ipDiffFOneLinearpfnStrongNonllRealX}{+0.026}
\newcommand{\ipDiffFOneMcThreeRealX}{\ensuremath{-0.050}}
\newcommand{\ipDiffFOneStabilityRefitRealX}{+0.219}
\newcommand{\ipDiffFOneSusieRealX}{+0.118}
\newcommand{\ipDiffHiAucExactRealX}{\ensuremath{-0.007}}
\newcommand{\ipDiffHiAucGlinternetRealX}{+0.176}
\newcommand{\ipDiffHiAucHierNetRealX}{+0.046}
\newcommand{\ipDiffHiAucLassoRealX}{+0.146}

\newcommand{\ipDiffHiAucMcThreeRealX}{\ensuremath{-0.005}}
\newcommand{\ipDiffHiAucStabilityRefitRealX}{+0.150}
\newcommand{\ipDiffHiAucSusieRealX}{+0.113}
\newcommand{\ipDiffHiFOneExactRealX}{\ensuremath{-0.024}}
\newcommand{\ipDiffHiFOneGlinternetRealX}{+0.210}
\newcommand{\ipDiffHiFOneHierNetRealX}{+0.249}
\newcommand{\ipDiffHiFOneLassoRealX}{+0.298}

\newcommand{\ipDiffHiFOneMcThreeRealX}{\ensuremath{-0.036}}
\newcommand{\ipDiffHiFOneStabilityRefitRealX}{+0.223}
\newcommand{\ipDiffHiFOneSusieRealX}{+0.122}
\newcommand{\ipDiffHiRmseExactRealX}{+0.017}
\newcommand{\ipDiffHiRmseGlinternetRealX}{\ensuremath{-0.003}}
\newcommand{\ipDiffHiRmseHierNetRealX}{0.000}
\newcommand{\ipDiffHiRmseLassoRealX}{\ensuremath{-0.007}}

\newcommand{\ipDiffHiRmseMcThreeRealX}{+0.012}
\newcommand{\ipDiffHiRmseStabilityRefitRealX}{\ensuremath{-0.019}}
\newcommand{\ipDiffHiRmseSusieRealX}{\ensuremath{-0.010}}
\newcommand{\ipDiffLoAucExactRealX}{\ensuremath{-0.012}}
\newcommand{\ipDiffLoAucGlinternetRealX}{+0.170}
\newcommand{\ipDiffLoAucHierNetRealX}{+0.043}
\newcommand{\ipDiffLoAucLassoRealX}{+0.141}

\newcommand{\ipDiffLoAucMcThreeRealX}{\ensuremath{-0.012}}
\newcommand{\ipDiffLoAucStabilityRefitRealX}{+0.144}
\newcommand{\ipDiffLoAucSusieRealX}{+0.107}
\newcommand{\ipDiffLoFOneExactRealX}{\ensuremath{-0.035}}
\newcommand{\ipDiffLoFOneGlinternetRealX}{+0.200}
\newcommand{\ipDiffLoFOneHierNetRealX}{+0.238}
\newcommand{\ipDiffLoFOneLassoRealX}{+0.288}

\newcommand{\ipDiffLoFOneMcThreeRealX}{\ensuremath{-0.064}}
\newcommand{\ipDiffLoFOneStabilityRefitRealX}{+0.214}
\newcommand{\ipDiffLoFOneSusieRealX}{+0.114}
\newcommand{\ipDiffLoRmseExactRealX}{+0.014}
\newcommand{\ipDiffLoRmseGlinternetRealX}{\ensuremath{-0.005}}
\newcommand{\ipDiffLoRmseHierNetRealX}{\ensuremath{-0.002}}
\newcommand{\ipDiffLoRmseLassoRealX}{\ensuremath{-0.009}}

\newcommand{\ipDiffLoRmseMcThreeRealX}{+0.007}
\newcommand{\ipDiffLoRmseStabilityRefitRealX}{\ensuremath{-0.043}}
\newcommand{\ipDiffLoRmseSusieRealX}{\ensuremath{-0.012}}
\newcommand{\ipDiffRmseExactRealX}{+0.015}
\newcommand{\ipDiffRmseGlinternetRealX}{\ensuremath{-0.004}}
\newcommand{\ipDiffRmseHierNetRealX}{\ensuremath{-0.001}}
\newcommand{\ipDiffRmseLassoRealX}{\ensuremath{-0.008}}

\newcommand{\ipDiffRmseLinearpfnStrongNonllRealX}{\ensuremath{-0.005}}
\newcommand{\ipDiffRmseMcThreeRealX}{+0.009}
\newcommand{\ipDiffRmseStabilityRefitRealX}{\ensuremath{-0.027}}
\newcommand{\ipDiffRmseSusieRealX}{\ensuremath{-0.011}}

\newcommand{\ipFOneMeanGlinternetRealX}{0.57}
\newcommand{\ipFOneMeanHierNetRealX}{0.53}
\newcommand{\ipFOneMeanLassoRealX}{0.48}

\newcommand{\ipFOneMeanLinearpfnStrongNonllRealX}{0.75}

\newcommand{\ipFOneMeanOursRealX}{0.77}
\newcommand{\ipFOneMeanStabilityRefitRealX}{0.55}
\newcommand{\ipFOneMeanSusieRealX}{0.66}

\newcommand{\ipFOneQOneGlinternetRealX}{0.43}
\newcommand{\ipFOneQOneHierNetRealX}{0.38}
\newcommand{\ipFOneQOneLassoRealX}{0.33}

\newcommand{\ipFOneQOneOursRealX}{0.67}
\newcommand{\ipFOneQOneStabilityRefitRealX}{0.40}
\newcommand{\ipFOneQOneSusieRealX}{0.46}

\newcommand{\ipFOneQThreeGlinternetRealX}{0.71}
\newcommand{\ipFOneQThreeHierNetRealX}{0.67}
\newcommand{\ipFOneQThreeLassoRealX}{0.63}

\newcommand{\ipFOneQThreeOursRealX}{1.00}
\newcommand{\ipFOneQThreeStabilityRefitRealX}{0.75}
\newcommand{\ipFOneQThreeSusieRealX}{0.89}

\newcommand{\ipFitMedExactRealX}{0.04}
\newcommand{\ipFitMedGlinternetRealX}{4.39}
\newcommand{\ipFitMedHierNetRealX}{25.45}
\newcommand{\ipFitMedLassoRealX}{0.14}

\newcommand{\ipFitMedMcThreeRealX}{57.43}
\newcommand{\ipFitMedOursRealX}{7.40}
\newcommand{\ipFitMedStabilityRefitRealX}{0.36}
\newcommand{\ipFitMedSusieRealX}{0.16}

\newcommand{\ipFitPNinetyMcThreeRealX}{866.31}

\newcommand{\ipFitQOneGlinternetRealX}{1.18}
\newcommand{\ipFitQOneHierNetRealX}{9.17}
\newcommand{\ipFitQOneLassoRealX}{0.08}

\newcommand{\ipFitQOneOursRealX}{1.07}
\newcommand{\ipFitQOneStabilityRefitRealX}{0.33}
\newcommand{\ipFitQOneSusieRealX}{0.08}

\newcommand{\ipFitQThreeGlinternetRealX}{15.05}
\newcommand{\ipFitQThreeHierNetRealX}{79.28}
\newcommand{\ipFitQThreeLassoRealX}{0.71}

\newcommand{\ipFitQThreeOursRealX}{24.59}
\newcommand{\ipFitQThreeStabilityRefitRealX}{0.49}
\newcommand{\ipFitQThreeSusieRealX}{0.40}

\newcommand{\ipFitRatioMcThreeRealX}{9.1}

\newcommand{\ipHzDiffHiRmseGlinternetRealX}{\ensuremath{-0.002}}
\newcommand{\ipHzDiffHiRmseHierNetRealX}{+0.001}
\newcommand{\ipHzDiffHiRmseLassoRealX}{\ensuremath{-0.006}}
\newcommand{\ipHzDiffHiRmseStabilityRefitRealX}{\ensuremath{-0.018}}
\newcommand{\ipHzDiffHiRmseSusieRealX}{\ensuremath{-0.009}}
\newcommand{\ipHzDiffLoRmseGlinternetRealX}{\ensuremath{-0.004}}
\newcommand{\ipHzDiffLoRmseHierNetRealX}{\ensuremath{-0.001}}
\newcommand{\ipHzDiffLoRmseLassoRealX}{\ensuremath{-0.008}}
\newcommand{\ipHzDiffLoRmseStabilityRefitRealX}{\ensuremath{-0.042}}
\newcommand{\ipHzDiffLoRmseSusieRealX}{\ensuremath{-0.011}}
\newcommand{\ipHzDiffRmseGlinternetRealX}{\ensuremath{-0.003}}
\newcommand{\ipHzDiffRmseHierNetRealX}{0.000}
\newcommand{\ipHzDiffRmseLassoRealX}{\ensuremath{-0.007}}
\newcommand{\ipHzDiffRmseStabilityRefitRealX}{\ensuremath{-0.026}}
\newcommand{\ipHzDiffRmseSusieRealX}{\ensuremath{-0.010}}
\newcommand{\ipHzPRmseGlinternetRealX}{= 0.775}

\newcommand{\ipIntCountMeanRealX}{3.50}
\newcommand{\ipIntCountMedianRealX}{1}
\newcommand{\ipIntNoneShareRealX}{37.5\%}
\newcommand{\ipLevelHiRealX}{0.9}
\newcommand{\ipLevelLoRealX}{0.1}
\newcommand{\ipLevelsRealX}{5}

\newcommand{\ipMarginFOneNMaxBinRealX}{200--499}

\newcommand{\ipMcmcConvPctRealX}{93.0\%}
\newcommand{\ipMcmcEssMedRealX}{4695}
\newcommand{\ipMcmcNRealX}{300}
\newcommand{\ipMcmcRhatMedRealX}{1.001}
\newcommand{\ipMcmcStepsMedRealX}{200,000}

\newcommand{\ipPAucLinearpfnStrongNonllRealX}{< 0.001}

\newcommand{\ipPFOneLinearpfnStrongNonllRealX}{< 0.001}

\newcommand{\ipPRmseHierNetRealX}{= 0.155}

\newcommand{\ipPRmseLinearpfnStrongNonllRealX}{< 0.001}

\newcommand{\ipPairedNFOneMcThreeRealX}{300}

\newcommand{\ipPerCellRealX}{400}

\newcommand{\ipRmseMeanExactRealX}{0.050}
\newcommand{\ipRmseMeanGlinternetRealX}{0.044}
\newcommand{\ipRmseMeanHierNetRealX}{0.042}
\newcommand{\ipRmseMeanLassoRealX}{0.049}

\newcommand{\ipRmseMeanOursRealX}{0.041}
\newcommand{\ipRmseMeanStabilityRefitRealX}{0.068}
\newcommand{\ipRmseMeanSusieRealX}{0.051}

\newcommand{\ipRmseQOneGlinternetRealX}{0.011}
\newcommand{\ipRmseQOneHierNetRealX}{0.011}
\newcommand{\ipRmseQOneLassoRealX}{0.012}

\newcommand{\ipRmseQOneOursRealX}{0.011}
\newcommand{\ipRmseQOneStabilityRefitRealX}{0.014}
\newcommand{\ipRmseQOneSusieRealX}{0.009}

\newcommand{\ipRmseQThreeGlinternetRealX}{0.052}
\newcommand{\ipRmseQThreeHierNetRealX}{0.051}
\newcommand{\ipRmseQThreeLassoRealX}{0.059}

\newcommand{\ipRmseQThreeOursRealX}{0.051}
\newcommand{\ipRmseQThreeStabilityRefitRealX}{0.074}
\newcommand{\ipRmseQThreeSusieRealX}{0.066}

\newcommand{\ipSupportNRealX}{10,000}

\newcommand{\ipUnitDiffFOneHierNetCrossSectionRealX}{+0.244}
\newcommand{\ipUnitDiffFOneHierNetPanelRealX}{+0.241}

\newcommand{\ipUnitDiffFOneSusieCrossSectionRealX}{+0.120}
\newcommand{\ipUnitDiffFOneSusiePanelRealX}{+0.110}

\newcommand{\ipUnitFOneMeanOursCrossSectionRealX}{0.77}
\newcommand{\ipUnitFOneMeanOursPanelRealX}{0.78}

\newcommand{\ipUnitNPanelRealX}{1,645}

\newcommand{\ipWinRmseLinearpfnStrongNonllRealX}{67.0\%}

\newcommand{\knobAucGlinternetCoef}{0.85}
\newcommand{\knobAucGlinternetInt}{0.75}
\newcommand{\knobAucGlinternetNoise}{0.80}
\newcommand{\knobAucGlinternetNone}{0.79}
\newcommand{\knobAucHierNetCoef}{0.94}
\newcommand{\knobAucHierNetInt}{0.88}
\newcommand{\knobAucHierNetNoise}{0.92}
\newcommand{\knobAucHierNetNone}{0.92}
\newcommand{\knobAucLassoCoef}{0.86}
\newcommand{\knobAucLassoInt}{0.81}
\newcommand{\knobAucLassoNoise}{0.82}
\newcommand{\knobAucLassoNone}{0.82}

\newcommand{\knobAucOursCoef}{0.98}
\newcommand{\knobAucOursInt}{0.94}
\newcommand{\knobAucOursNoise}{0.97}
\newcommand{\knobAucOursNone}{0.96}

\newcommand{\knobAucStabilityRefitCoef}{0.86}
\newcommand{\knobAucStabilityRefitInt}{0.80}
\newcommand{\knobAucStabilityRefitNoise}{0.82}
\newcommand{\knobAucStabilityRefitNone}{0.82}
\newcommand{\knobAucSusieCoef}{0.89}
\newcommand{\knobAucSusieInt}{0.83}
\newcommand{\knobAucSusieNoise}{0.86}
\newcommand{\knobAucSusieNone}{0.85}

\newcommand{\knobCtrlDiffAucGlinternetCoef}{+0.062}
\newcommand{\knobCtrlDiffAucGlinternetInt}{\ensuremath{-0.045}}
\newcommand{\knobCtrlDiffAucGlinternetNoise}{+0.005}
\newcommand{\knobCtrlDiffAucHierNetCoef}{+0.018}
\newcommand{\knobCtrlDiffAucHierNetInt}{\ensuremath{-0.039}}
\newcommand{\knobCtrlDiffAucHierNetNoise}{+0.002}
\newcommand{\knobCtrlDiffAucLassoCoef}{+0.040}
\newcommand{\knobCtrlDiffAucLassoInt}{\ensuremath{-0.009}}
\newcommand{\knobCtrlDiffAucLassoNoise}{+0.003}

\newcommand{\knobCtrlDiffAucOursCoef}{+0.011}
\newcommand{\knobCtrlDiffAucOursInt}{\ensuremath{-0.026}}
\newcommand{\knobCtrlDiffAucOursNoise}{+0.002}
\newcommand{\knobCtrlDiffAucStabilityRefitCoef}{+0.042}
\newcommand{\knobCtrlDiffAucStabilityRefitInt}{\ensuremath{-0.015}}
\newcommand{\knobCtrlDiffAucStabilityRefitNoise}{+0.003}
\newcommand{\knobCtrlDiffAucSusieCoef}{+0.034}
\newcommand{\knobCtrlDiffAucSusieInt}{\ensuremath{-0.021}}
\newcommand{\knobCtrlDiffAucSusieNoise}{+0.002}

\newcommand{\knobCtrlDiffFOneGlinternetCoef}{\ensuremath{-0.011}}
\newcommand{\knobCtrlDiffFOneGlinternetInt}{+0.029}
\newcommand{\knobCtrlDiffFOneGlinternetNoise}{+0.006}
\newcommand{\knobCtrlDiffFOneHierNetCoef}{\ensuremath{-0.006}}
\newcommand{\knobCtrlDiffFOneHierNetInt}{+0.034}
\newcommand{\knobCtrlDiffFOneHierNetNoise}{+0.005}
\newcommand{\knobCtrlDiffFOneLassoCoef}{+0.004}
\newcommand{\knobCtrlDiffFOneLassoInt}{+0.044}
\newcommand{\knobCtrlDiffFOneLassoNoise}{+0.005}

\newcommand{\knobCtrlDiffFOneOursCoef}{+0.072}
\newcommand{\knobCtrlDiffFOneOursInt}{\ensuremath{-0.017}}
\newcommand{\knobCtrlDiffFOneOursNoise}{+0.014}
\newcommand{\knobCtrlDiffFOneStabilityRefitCoef}{+0.046}
\newcommand{\knobCtrlDiffFOneStabilityRefitInt}{0.000}
\newcommand{\knobCtrlDiffFOneStabilityRefitNoise}{+0.005}
\newcommand{\knobCtrlDiffFOneSusieCoef}{+0.044}
\newcommand{\knobCtrlDiffFOneSusieInt}{\ensuremath{-0.014}}
\newcommand{\knobCtrlDiffFOneSusieNoise}{+0.007}

\newcommand{\knobCtrlDiffHiRmseOursInt}{+0.024}
\newcommand{\knobCtrlDiffHiRmseOursNoise}{\ensuremath{-0.001}}

\newcommand{\knobCtrlDiffLoRmseOursInt}{+0.022}
\newcommand{\knobCtrlDiffLoRmseOursNoise}{\ensuremath{-0.001}}

\newcommand{\knobCtrlDiffRmseGlinternetCoef}{+0.004}
\newcommand{\knobCtrlDiffRmseGlinternetInt}{+0.015}
\newcommand{\knobCtrlDiffRmseGlinternetNoise}{\ensuremath{-0.001}}
\newcommand{\knobCtrlDiffRmseHierNetCoef}{+0.003}
\newcommand{\knobCtrlDiffRmseHierNetInt}{+0.018}
\newcommand{\knobCtrlDiffRmseHierNetNoise}{\ensuremath{-0.002}}
\newcommand{\knobCtrlDiffRmseLassoCoef}{+0.007}
\newcommand{\knobCtrlDiffRmseLassoInt}{+0.012}
\newcommand{\knobCtrlDiffRmseLassoNoise}{\ensuremath{-0.002}}

\newcommand{\knobCtrlDiffRmseOursCoef}{+0.003}
\newcommand{\knobCtrlDiffRmseOursInt}{+0.023}
\newcommand{\knobCtrlDiffRmseOursNoise}{\ensuremath{-0.001}}
\newcommand{\knobCtrlDiffRmseStabilityRefitCoef}{+0.030}
\newcommand{\knobCtrlDiffRmseStabilityRefitInt}{+0.050}
\newcommand{\knobCtrlDiffRmseStabilityRefitNoise}{+0.123}
\newcommand{\knobCtrlDiffRmseSusieCoef}{+0.012}
\newcommand{\knobCtrlDiffRmseSusieInt}{+0.021}
\newcommand{\knobCtrlDiffRmseSusieNoise}{\ensuremath{-0.001}}

\newcommand{\knobDatasetsOursCoef}{10,000}

\newcommand{\knobDiffFOneSusieInt}{+0.115}

\newcommand{\knobDiffHiFOneSusieInt}{+0.119}

\newcommand{\knobDiffHiRmseGlinternetInt}{+0.006}

\newcommand{\knobDiffLoFOneSusieInt}{+0.112}

\newcommand{\knobDiffLoRmseGlinternetInt}{+0.003}

\newcommand{\knobDiffRmseGlinternetInt}{+0.005}

\newcommand{\knobEceOursCoef}{0.008}
\newcommand{\knobEceOursInt}{0.026}
\newcommand{\knobEceOursNoise}{0.007}
\newcommand{\knobEceOursNone}{0.009}

\newcommand{\knobEceSusieNone}{0.041}

\newcommand{\knobExactEceExactCoef}{0.028}
\newcommand{\knobExactEceExactInt}{0.092}

\newcommand{\knobExactEceExactNone}{0.063}

\newcommand{\knobExactEceOursCoef}{0.036}
\newcommand{\knobExactEceOursInt}{0.124}

\newcommand{\knobExactEceOursNone}{0.081}

\newcommand{\knobExactFOneExactInt}{0.82}

\newcommand{\knobExactFOneExactNone}{0.82}

\newcommand{\knobExactFOneOursInt}{0.78}

\newcommand{\knobExactFOneOursNone}{0.79}

\newcommand{\knobFOneGlinternetCoef}{0.56}
\newcommand{\knobFOneGlinternetInt}{0.60}
\newcommand{\knobFOneGlinternetNoise}{0.57}
\newcommand{\knobFOneGlinternetNone}{0.57}
\newcommand{\knobFOneHierNetCoef}{0.52}
\newcommand{\knobFOneHierNetInt}{0.56}
\newcommand{\knobFOneHierNetNoise}{0.53}
\newcommand{\knobFOneHierNetNone}{0.53}
\newcommand{\knobFOneLassoCoef}{0.48}
\newcommand{\knobFOneLassoInt}{0.52}
\newcommand{\knobFOneLassoNoise}{0.49}
\newcommand{\knobFOneLassoNone}{0.48}

\newcommand{\knobFOneOursCoef}{0.85}
\newcommand{\knobFOneOursInt}{0.76}
\newcommand{\knobFOneOursNoise}{0.79}
\newcommand{\knobFOneOursNone}{0.77}

\newcommand{\knobFOneStabilityRefitCoef}{0.60}
\newcommand{\knobFOneStabilityRefitInt}{0.55}
\newcommand{\knobFOneStabilityRefitNoise}{0.56}
\newcommand{\knobFOneStabilityRefitNone}{0.55}
\newcommand{\knobFOneSusieCoef}{0.70}
\newcommand{\knobFOneSusieInt}{0.64}
\newcommand{\knobFOneSusieNoise}{0.66}
\newcommand{\knobFOneSusieNone}{0.66}

\newcommand{\knobIntCountMaxInt}{135}

\newcommand{\knobIntCountMeanInt}{9.09}

\newcommand{\knobIntCountMedianInt}{3}

\newcommand{\knobIntNoneShareInt}{15.7\%}

\newcommand{\knobRmseGlinternetCoef}{0.048}
\newcommand{\knobRmseGlinternetInt}{0.059}
\newcommand{\knobRmseGlinternetNoise}{0.043}
\newcommand{\knobRmseGlinternetNone}{0.044}
\newcommand{\knobRmseHierNetCoef}{0.045}
\newcommand{\knobRmseHierNetInt}{0.060}
\newcommand{\knobRmseHierNetNoise}{0.040}
\newcommand{\knobRmseHierNetNone}{0.042}
\newcommand{\knobRmseLassoCoef}{0.056}
\newcommand{\knobRmseLassoInt}{0.061}
\newcommand{\knobRmseLassoNoise}{0.047}
\newcommand{\knobRmseLassoNone}{0.049}

\newcommand{\knobRmseOursCoef}{0.044}
\newcommand{\knobRmseOursInt}{0.064}
\newcommand{\knobRmseOursNoise}{0.040}
\newcommand{\knobRmseOursNone}{0.041}
\newcommand{\knobRmseStabilityRefitCoef}{0.098}
\newcommand{\knobRmseStabilityRefitInt}{0.118}
\newcommand{\knobRmseStabilityRefitNoise}{0.191}
\newcommand{\knobRmseStabilityRefitNone}{0.068}
\newcommand{\knobRmseSusieCoef}{0.063}
\newcommand{\knobRmseSusieInt}{0.072}
\newcommand{\knobRmseSusieNoise}{0.051}
\newcommand{\knobRmseSusieNone}{0.051}

\newcommand{\modelBarHi}{8}
\newcommand{\modelBarLo}{-8}
\newcommand{\modelBins}{512}
\newcommand{\modelEmb}{512}
\newcommand{\modelHeads}{16}
\newcommand{\modelLayers}{8}
\newcommand{\modelMlp}{1,024}
\newcommand{\modelParams}{31.5M}
\newcommand{\modelParamsExact}{31,541,766}

\newcommand{\priorFactorSharePFiveZero}{0.25}

\newcommand{\priorFactorShareZero}{19.7\%}
\newcommand{\priorGridBenchCellsTotalPct}{19.1}

\newcommand{\priorGridDatasets}{60,000}

\newcommand{\priorIntPFiveZero}{0.099}

\newcommand{\priorIntPSevenFive}{0.172}
\newcommand{\priorIntPTwoFive}{0.046}

\newcommand{\priorMainsPFiveZero}{0.169}

\newcommand{\priorMainsPSevenFive}{0.294}
\newcommand{\priorMainsPTwoFive}{0.078}
\newcommand{\priorNDraws}{10,000}
\newcommand{\priorNormPFiveZero}{0.19}
\newcommand{\priorNormPSevenFive}{0.29}
\newcommand{\priorNormPTwoFive}{0.11}
\newcommand{\priorNullFrac}{0.103}
\newcommand{\priorNullFracPTwo}{0.273}

\newcommand{\probeJitterSd}{0.1}
\newcommand{\probePointsPerEffect}{3}
\newcommand{\probeRidge}{\ensuremath{1.0\times10^{-3}}}

\newcommand{\quadByCMaxDPipIntOneTwo}{\ensuremath{4.1\times10^{-4}}}

\newcommand{\quadCGridInt}{12, 15 and 20}
\newcommand{\quadCellsNInt}{32 and 1,024}
\newcommand{\quadCellsPInt}{3 and 5}
\newcommand{\quadChosenMaxDLogEvInt}{\ensuremath{9.8\times10^{-6}}}
\newcommand{\quadChosenMaxDPipInt}{\ensuremath{1.4\times10^{-7}}}
\newcommand{\quadChosenWorstNInt}{1,024}
\newcommand{\quadChosenWorstPInt}{5}
\newcommand{\quadFinestCInt}{30}

\newcommand{\quadNDatasets}{12}
\newcommand{\quadNodesC}{20}

\newcommand{\quadPruneTol}{\ensuremath{1.0\times10^{-9}}}

\newcommand{\realyChains}{4}
\newcommand{\realyCoefRHeadAll}{0.991}

\newcommand{\realyCoefRmseHeadAll}{0.014}
\newcommand{\realyCoefRmseHeadAttitude}{0.006}
\newcommand{\realyCoefRmseHeadBodyfat}{0.019}
\newcommand{\realyCoefRmseHeadDiabetes}{0.011}

\newcommand{\realyCoefRmseHeadOzone}{0.013}
\newcommand{\realyCoefRmseHeadProstate}{0.009}
\newcommand{\realyCoefRmseHeadSwiss}{0.019}
\newcommand{\realyCoefRmseHeadTalOr}{0.009}

\newcommand{\realyEssMin}{899}

\newcommand{\realyEssThreshold}{200}

\newcommand{\realyMpmIdentical}{5}

\newcommand{\realyMpmSameAttitude}{21}
\newcommand{\realyMpmSameBodyfat}{87}
\newcommand{\realyMpmSameDiabetes}{53}
\newcommand{\realyMpmSameOzone}{45}
\newcommand{\realyMpmSameProstate}{36}
\newcommand{\realyMpmSameSwiss}{15}
\newcommand{\realyMpmSameTalOr}{15}

\newcommand{\realyMpmSlots}{278}
\newcommand{\realyMpmSlotsAttitude}{21}
\newcommand{\realyMpmSlotsBodyfat}{91}
\newcommand{\realyMpmSlotsDiabetes}{55}
\newcommand{\realyMpmSlotsOzone}{45}
\newcommand{\realyMpmSlotsProstate}{36}
\newcommand{\realyMpmSlotsSame}{272}

\newcommand{\realyMpmSlotsSwiss}{15}
\newcommand{\realyMpmSlotsTalOr}{15}
\newcommand{\realyN}{7}
\newcommand{\realyNAttitude}{30}
\newcommand{\realyNBodyfat}{252}
\newcommand{\realyNDiabetes}{442}
\newcommand{\realyNMax}{442}
\newcommand{\realyNMin}{30}
\newcommand{\realyNOzone}{330}
\newcommand{\realyNProstate}{97}
\newcommand{\realyNStress}{3}
\newcommand{\realyNSwiss}{47}
\newcommand{\realyNTalOr}{123}

\newcommand{\realyNllMarginal}{1.385}

\newcommand{\realyNllOlsAttitude}{1.177}
\newcommand{\realyNllOlsBodyfat}{0.667}
\newcommand{\realyNllOlsDiabetes}{1.039}
\newcommand{\realyNllOlsOzone}{0.839}
\newcommand{\realyNllOlsProstate}{0.876}
\newcommand{\realyNllOlsSwiss}{1.765}
\newcommand{\realyNllOlsTalOr}{1.246}

\newcommand{\realyNllOursAttitude}{1.095}
\newcommand{\realyNllOursBodyfat}{0.651}
\newcommand{\realyNllOursDiabetes}{1.037}
\newcommand{\realyNllOursOzone}{0.742}
\newcommand{\realyNllOursProstate}{0.927}
\newcommand{\realyNllOursSwiss}{1.277}
\newcommand{\realyNllOursTalOr}{1.226}

\newcommand{\realyNllRefAttitude}{1.122}
\newcommand{\realyNllRefBodyfat}{0.663}
\newcommand{\realyNllRefDiabetes}{1.027}
\newcommand{\realyNllRefOzone}{0.785}
\newcommand{\realyNllRefProstate}{0.894}
\newcommand{\realyNllRefSwiss}{1.312}
\newcommand{\realyNllRefTalOr}{1.226}

\newcommand{\realyPAttitude}{6}
\newcommand{\realyPBodyfat}{13}
\newcommand{\realyPDiabetes}{10}
\newcommand{\realyPMax}{13}
\newcommand{\realyPMin}{5}
\newcommand{\realyPOzone}{9}
\newcommand{\realyPProstate}{8}
\newcommand{\realyPSwiss}{5}
\newcommand{\realyPTalOr}{5}

\newcommand{\realyPipMaxAbsAllAttitude}{0.043}
\newcommand{\realyPipMaxAbsAllBodyfat}{0.600}
\newcommand{\realyPipMaxAbsAllDiabetes}{0.248}
\newcommand{\realyPipMaxAbsAllOzone}{0.213}
\newcommand{\realyPipMaxAbsAllProstate}{0.107}
\newcommand{\realyPipMaxAbsAllSwiss}{0.157}
\newcommand{\realyPipMaxAbsAllTalOr}{0.044}

\newcommand{\realyPipRAll}{0.972}

\newcommand{\realyPipRInts}{0.917}

\newcommand{\realyPipRMains}{0.937}
\newcommand{\realyPipRMainsAttitude}{0.998}
\newcommand{\realyPipRMainsBodyfat}{0.780}
\newcommand{\realyPipRMainsDiabetes}{0.961}
\newcommand{\realyPipRMainsOzone}{0.966}
\newcommand{\realyPipRMainsProstate}{0.990}
\newcommand{\realyPipRMainsSwiss}{0.854}
\newcommand{\realyPipRMainsTalOr}{0.999}
\newcommand{\realyPipRmseAll}{0.065}
\newcommand{\realyPipRmseAllAttitude}{0.016}
\newcommand{\realyPipRmseAllBodyfat}{0.083}
\newcommand{\realyPipRmseAllDiabetes}{0.061}
\newcommand{\realyPipRmseAllOzone}{0.069}
\newcommand{\realyPipRmseAllProstate}{0.028}
\newcommand{\realyPipRmseAllSwiss}{0.074}
\newcommand{\realyPipRmseAllTalOr}{0.017}

\newcommand{\realyQuerySharePct}{19.9\%}

\newcommand{\realyRhatMax}{1.006}

\newcommand{\realyRhatThreshold}{1.1}

\newcommand{\realyStepsMax}{500,000}
\newcommand{\realyStepsMin}{100,000}

\newcommand{\realyStressTables}{bodyfat, ozone and diabetes}

\newcommand{\rxEligibleN}{383}
\newcommand{\rxExclDescriptionKeyword}{17}
\newcommand{\rxExclDownloadFailed}{3}
\newcommand{\rxExclDuplicateColumns}{23}
\newcommand{\rxExclDuplicateTitle}{8}
\newcommand{\rxExclIndexPModelLtMin}{81}
\newcommand{\rxExclIndexRowsLtMin}{93}
\newcommand{\rxExclObjectClass}{69}
\newcommand{\rxExclPackageExcluded}{2,908}
\newcommand{\rxExclPureTimeSeries}{32}
\newcommand{\rxExclRealizedNLtMin}{3}
\newcommand{\rxExclRealizedPLtMin}{15}
\newcommand{\rxExclRowLossGtMax}{13}

\newcommand{\rxExclTotal}{3,265}

\newcommand{\rxFrameN}{480}

\newcommand{\rxIncludedPackages}{20}
\newcommand{\rxIndexN}{3,648}
\newcommand{\rxIndexPackages}{104}
\newcommand{\rxMaxAbsCor}{0.95}

\newcommand{\rxMaxColMissing}{50.0\%}
\newcommand{\rxMaxRowLoss}{50.0\%}

\newcommand{\rxMinMinorityRows}{10}
\newcommand{\rxMinN}{30}
\newcommand{\rxMinP}{3}

\newcommand{\rxPoolColSubsampled}{15}
\newcommand{\rxPoolDomainCriminology}{17}
\newcommand{\rxPoolDomainDemography}{1}
\newcommand{\rxPoolDomainEconomics}{237}
\newcommand{\rxPoolDomainEducation}{24}

\newcommand{\rxPoolDomainPoliticalScience}{58}
\newcommand{\rxPoolDomainPsychology}{14}
\newcommand{\rxPoolDomainSocialScience}{32}
\newcommand{\rxPoolDomainsN}{7}
\newcommand{\rxPoolMaxAbsRMax}{0.965}
\newcommand{\rxPoolMaxAbsRMedian}{0.80}
\newcommand{\rxPoolN}{383}
\newcommand{\rxPoolNMax}{1,999}
\newcommand{\rxPoolNMedian}{758}

\newcommand{\rxPoolPLeFive}{87}
\newcommand{\rxPoolPMax}{30}
\newcommand{\rxPoolPMedian}{9}

\newcommand{\rxPoolPackages}{20}

\newcommand{\rxPoolRowSubsampled}{121}

\newcommand{\rxPoolRowUnitPanel}{61}

\newcommand{\stabPairs}{50}
\newcommand{\stabThreshold}{0.75}
\newcommand{\susieL}{20}
\newcommand{\trainAccum}{16}
\newcommand{\trainCkptMinutes}{30}
\newcommand{\trainDatasetsSeen}{2,400,000}
\newcommand{\trainDatasetsSeenBig}{2.4M}
\newcommand{\trainEffBatch}{64}

\newcommand{\trainGpu}{H200}
\newcommand{\trainGpuCount}{2}
\newcommand{\trainGpuCountWord}{two}
\newcommand{\trainGpuHours}{29.1}

\newcommand{\trainLr}{0.001}
\newcommand{\trainMuonLr}{0.02}
\newcommand{\trainNMax}{2,000}
\newcommand{\trainNMin}{20}

\newcommand{\trainPMax}{30}
\newcommand{\trainPMin}{2}
\newcommand{\trainPeakMemGiB}{57.3}
\newcommand{\trainPhysBatch}{4}
\newcommand{\trainQueryFracHi}{0.5}
\newcommand{\trainQueryFracLo}{0.1}

\newcommand{\trainRestartsWord}{once}
\newcommand{\trainResumedStep}{5,684}
\newcommand{\trainSteps}{37,500}
\newcommand{\trainStepsPerSec}{0.72}

\newcommand{\trainWallHours}{14.6}

\newcommand{\trainWarmup}{4,000}
\newcommand{\valDatasetsPerCell}{100}
\newcommand{\valNQuery}{64}
\newcommand{\valNValues}{32, 128, 512, 1,024 and 2,000}
\newcommand{\valPValues}{3, 5, 8, 10, 15, 20 and 30}

\usepackage{hyperref}
\usepackage{url}
\usepackage{graphicx}
\usepackage{wrapfig}
\usepackage{xcolor}
\usepackage{csquotes}

\usepackage{acro}
\DeclareAcronym{pip}{short=PIP, long=posterior inclusion probability}
\DeclareAcronym{pfn}{short=PFN, long=prior-data fitted network}
\DeclareAcronym{mpm}{short=MPM, long=median probability model}
\DeclareAcronym{ece}{short=ECE, long=expected calibration error}
\DeclareAcronym{auc}{short=AUC, long=area under the ROC curve}
\DeclareAcronym{mcmc}{short=MCMC, long=Markov chain Monte Carlo}
\DeclareAcronym{bce}{short=BCE, long=binary cross-entropy}
\DeclareAcronym{ols}{short=OLS, long=ordinary least squares}
\DeclareAcronym{rmse}{short=RMSE, long=root mean squared error}
\DeclareAcronym{npe}{short=NPE, long=neural posterior estimation}

\title{LinearPFN: Amortized Variable Selection for Linear Models with Interactions}

\author{Louis Schiekiera$^{1,2,3}$, Max Zimmer$^{3,4}$, Christophe Roux$^{3,4}$, Manuel Arnold$^{1}$, \\
\textbf{Sebastian Pokutta$^{3,4}$ \& Fritz G\"unther$^{1}$} \\[4pt]
$^{1}$Institute of Psychology, Humboldt-Universit\"at zu Berlin, Berlin, Germany \\
$^{2}$Department of Education and Psychology, Freie Universit\"at Berlin, Berlin, Germany \\
$^{3}$Department for AI in Society, Science, and Technology, Zuse Institute Berlin, Germany \\
$^{4}$Institute of Mathematics, Technische Universit\"at Berlin, Germany \\
Correspondence to: Louis Schiekiera \texttt{<louis.schiekiera@hu-berlin.de>}
}

\newcommand{\iqr}[2]{{\scriptsize [#1, #2]}}

\iclrfinalcopy 
\begin{document}

\maketitle
\lhead{Preprint}


\begin{abstract}
  Spike-and-slab regression is a standard Bayesian formulation of variable selection: it returns a posterior distribution over which candidate effects are active rather than a single selected subset, so that every candidate effect carries an inclusion probability. 
  Its cost grows exponentially with the number of candidate effects, so the posterior can be enumerated exactly only when the number of predictors is small.
  Beyond that reach, the posterior has to be approximated, typically by Markov chain Monte Carlo over the model space, which requires a fresh run for every dataset and, within a fixed budget of steps, may fail to converge.
  We present \textsc{LinearPFN}, a prior-data fitted transformer network that amortizes spike-and-slab inference for linear models with main effects and pairwise interactions.
  The network is pretrained once on synthetic datasets, drawn from an explicitly specified prior, and a single forward pass over a new dataset returns posterior inclusion probabilities, posterior-mean coefficients and posterior predictive distributions with no per-dataset fitting.
  The prior is conjugate by design, so that the posterior for each fixed set of active effects has a closed form, and wherever the exact posterior is still computable by enumeration we verify the network's outputs against it.
  On real predictor matrices from published social-science datasets, with outcomes drawn from the prior so that the true active set is known, \textsc{LinearPFN} attains a higher per-dataset selection AUC and a higher $F_1$ under the median probability model rule than five classical baselines. The lead holds when the coefficients, the interactions or the noise depart from the prior.
  Code: \url{https://github.com/schiekiera/LinearPFN}. Trained model: \url{https://huggingface.co/schiekiera/LinearPFN}.
\end{abstract}


\section{Introduction}
\label{sec:intro}

A linear model explains an outcome $y$ as a weighted sum of predictors, with one coefficient per predictor that gives the change in $y$ per unit of that predictor. Despite the rise of nonlinear statistical modeling, quantitative claims in the sciences are still frequently stated in this form \citep{cohen2013applied}: the coefficients are interpretable, their statistical properties are well understood, and each one can be tested directly against the theory that motivated it. Linear models are often extended with pairwise interactions, which state that the effect of one predictor on $y$ depends on the level of a second predictor \citep{aiken1991multiple}.

Fitting a linear-plus-interaction model requires a decision about which predictors and which interactions to include. A standard Bayesian formulation for variable selection is spike-and-slab regression \citep{mitchell1988bayesian,george1993variable}. Its prior either sets each coefficient exactly to zero (the spike) or draws it from a Gaussian (the slab). Under this prior, each model is defined by its active set, the effects whose coefficients are nonzero and thus drawn from the slab, and the posterior is a distribution over these models. The posterior probability that an effect is active is its \ac{pip}. With $p$ predictors the candidate effects are an intercept, $p$ main effects and $\binom{p}{2}$ pairwise interactions, $d = 1 + p + \binom{p}{2}$ in total. With the intercept always included there are $2^{d-1}$ models. At $p = 10$ this is $2^{55} \approx 3.6 \times 10^{16}$ models. The exact posterior is a sum over all models, so in practice it can be computed only at small $p$. Beyond that reach, practice relies on \ac{mcmc} over the model space \citep{madigan1995bayesian}, adaptive sampling \citep{clyde2011bayesian} or variational approximations \citep{susieR1}. 

\Acp{pfn} \citep{müller2024transformersbayesianinference} approximate a posterior by a different route. A transformer is trained on datasets sampled from a prior. Each dataset is split into context rows, which the network sees with their outcomes, and query rows, whose outcomes it predicts. The loss is evaluated on the query rows, so that the in-context predictions approximate the posterior predictive distribution under that prior. After training, no per-dataset fitting takes place. \textsc{TabPFN} \citep{hollmann2025accurate,hollmann2022tabpfn} applies this idea to general tabular prediction with a broad prior over mostly nonlinear structural causal models. We instead restrict the prior to the model class that the variable-selection tradition in statistics already uses, linear main effects and pairwise interactions under a spike-and-slab prior \citep{mitchell1988bayesian}. The prior is conjugate, meaning that the posterior under each candidate model has a closed form. Therefore, for a small number of predictors the exact posterior can be computed by enumerating the models, and the outputs of the network are validated against it for up to \exactMaxP{} predictors.

\begin{figure}[t]
\begin{center}
\includegraphics[width=\linewidth]{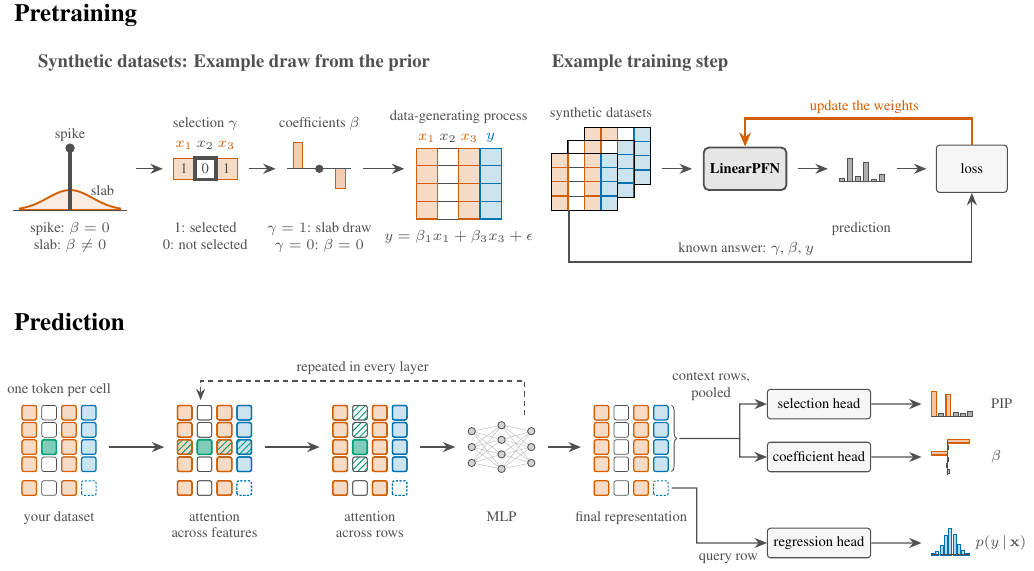}
\end{center}
\caption{Pretraining and prediction of \textsc{LinearPFN}. Top: each prior draw fixes an active set $\gamma$ and spike-or-slab coefficients $\beta$ for one synthetic dataset, and the loss compares the network's output with them. Bottom: in every layer, each cell (green) first attends to the other cells in its row and then to the other cells in its column (striped), and an MLP then updates each cell on its own. One forward pass over a new dataset returns inclusion probabilities and posterior-mean coefficients for every main effect and interaction, and a posterior predictive distribution for each query row.}

\label{fig:visual-abstract}
\end{figure}

\textsc{LinearPFN} (Figure~\ref{fig:visual-abstract}) is a transformer, based on \textsc{nanoTabPFN} \citep{pfefferle2025nanotabpfn}, pretrained on synthetic datasets drawn from an explicitly specified spike-and-slab prior. In one deterministic forward pass, a regression head returns the posterior predictive distribution, a selection head, trained with \ac{bce} against the simulated active sets in the manner of amortized causal discovery \citep{lorch2022amortized}, returns a \ac{pip} for every main effect and pairwise interaction, and a coefficient head returns posterior-mean coefficients. Our contributions are as follows:
\begin{enumerate}
\item \textbf{The first \ac{pfn} for Bayesian variable selection.} To our knowledge, no \ac{pfn} has returned inclusion probabilities over a regression model space. 
\item \textbf{Validation against the exact posterior.} Because the posterior under each candidate model has a closed form, the exact posterior can be enumerated wherever the number of candidate effects allows it, and every output of the network is scored against it on six criteria fixed before training.
\item \textbf{A prior calibrated to social-science data.} The pretraining prior is built to resemble individual-level data as collected in the social sciences. Effect sizes are calibrated so that the standardized effect of an active main effect matches meta-analytic norms for published correlations in psychology.
\item \textbf{Benchmark evaluation.} Practical performance is measured against five baselines on \ipDatasetsOursRealX{} datasets that combine real predictor matrices from social-science data with outcomes drawn from the prior. Three variants each violate one assumption of the outcome model. On published tables with their actual outcomes, \textsc{LinearPFN} is compared with an \ac{mcmc} sampler over the model space that targets the same posterior.
\end{enumerate}


\section{Related work}
\label{sec:related}


\textbf{Amortized Bayesian inference.}
\citet{müller2024transformersbayesianinference} demonstrated that \acp{pfn} approximate the posterior predictive distribution under a prior for Bayesian linear regression. \textsc{TabPFN} \citep{hollmann2022tabpfn,hollmann2025accurate} scaled the idea to general tabular prediction with a broad prior over mostly nonlinear structural causal models. \textsc{LinearPFN} is forked from \textsc{nanoTabPFN} \citep{pfefferle2025nanotabpfn}, a lightweight implementation of \textsc{TabPFN}. \Ac{npe} targets the parameters instead, usually with a normalizing flow over their joint posterior \citep{radev2022bayesflow,kipnis2025metabeta}. \textsc{LinearPFN} combines both: its regression head is trained as in a \ac{pfn}, and its selection and coefficient heads are supervised by the simulated active set and coefficients, as in \ac{npe}.



\textbf{Variable selection.}
Frequentist variable selection returns one subset of predictors, Bayesian selection a distribution over subsets. The lasso \citep{tibshirani1996regression} selects through an $\ell_1$ penalty that sets some coefficients exactly to zero, and stability selection \citep{meinshausenStabilitySelection2010} keeps the effects the lasso selects often under resampling. Applied to pairwise interactions, the lasso can keep an interaction and drop its parent main effects. \textsc{glinternet} \citep{gli2026} and \textsc{hierNet} \citep{hier2020} enforce heredity inside the penalty, through overlapping groups and a convex constraint, respectively. Under a spike-and-slab prior \citep{mitchell1988bayesian,george1993variable}, every candidate effect carries an inclusion probability, and coefficients are averaged over models. Samplers over the model space \citep{madigan1995bayesian,clyde2011bayesian} need a fitting run per dataset, and \textsc{SuSiE} \citep{susieR1,susieR2} approximates the posterior variationally under an assumed maximum number of effects. \citet{alrawkan2026linear} train a network on simulated linear models to classify predictors as active from \ac{ols} summary statistics, without inclusion probabilities.


\section{Methods}
\label{sec:methods}


\subsection{The prior}
\label{sec:prior}

\textsc{LinearPFN} is trained on synthetic datasets drawn from a prior over data-generating processes. A dataset has $n$ rows and $p$ predictors. For row $i$, the outcome follows a linear model with pairwise interactions, $y_i = \beta_0 + \sum_j \beta_j x_{ij} + \sum_{j<k} \beta_{jk}\, x_{ij} x_{ik} + \epsilon_i$, with the intercept $\beta_0$, the main-effect coefficients $\beta_j$ and the interaction coefficients $\beta_{jk}$. We write $e$ for any of the labels $0$, $j$ and $jk$, so $e$ runs over all $d = 1 + p + \tbinom{p}{2}$ candidate effects, and collect the coefficients in $\beta \in \R^d$. Given which effects are active, the coefficients follow a spike-and-slab prior, and the model reads in matrix form
\begin{equation}
\beta_e \mid \gamma_e, c, \sigma^2 \;\sim\; (1-\gamma_e)\,\delta_0 + \gamma_e\,\gN(0,\, v_e\,\sigma^2), \qquad
\mathbf{y} = \mathbf{Z}\beta + \epsilon,\quad \epsilon \sim \gN(0, \sigma^2 \mathbf{I}_n).
\label{eq:model}
\end{equation}
The expanded predictor matrix $\mathbf{Z} = [\mathbf{1},\ \mathbf{X},\ (\mathbf{x}_j\odot\mathbf{x}_k)_{j<k}] \in \R^{n\times d}$ has one column per candidate effect: a column of ones for the intercept, the $p$ z-standardized columns $\mathbf{x}_j$ of the predictor matrix $\mathbf{X} \in \R^{n\times p}$ for the main effects, and their elementwise products $\mathbf{x}_j\odot\mathbf{x}_k$ for the interactions. The active set $\gamma \in \{0,1\}^d$ marks which effects are active, with the intercept always active. An inactive effect ($\gamma_e = 0$) has a coefficient of zero, the point mass $\delta_0$ (spike). An active effect ($\gamma_e = 1$) has a Gaussian coefficient (slab) with variance $v_e\sigma^2$. The noise $\epsilon$ has variance $\sigma^2$. Three quantities are drawn once per dataset:

\begin{equation}
\gamma \sim \pi(\gamma \mid p),\qquad
c \sim \mathrm{LogNormal}(\mu_c, s_c^2),\qquad
\sigma^2 \sim \mathrm{InvGamma}(a_0, b_0).
\label{eq:hyper}
\end{equation}

\begin{figure}[t]
\begin{center}
\includegraphics[width=\linewidth]{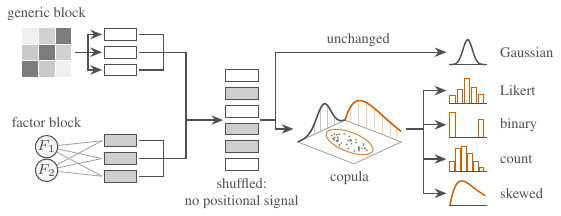}
\end{center}
\caption{Predictor prior. Left: the generic block is generated jointly from a random correlation matrix $R$ (top), and the factor block loads on latent factors (bottom). Middle: the two blocks are merged and the column order is shuffled. Right: a Gaussian copula maps each column to its marginal, either unchanged (Gaussian) or transformed to a Likert, binary, count or skewed family.}
\label{fig:predictor-prior}
\end{figure}

The active-effects prior $\pi(\gamma \mid p)$ is a distribution over active sets, described below. The slab scale $c$ (truncated to $[c_{\min}, c_{\max}]$) sets how large the effects are, and $\sigma^2$ sets the noise level. The slab variance is $v_e = c$ for main effects, $v_e = c\rho$ for interactions, and $v_e = \tau_0^2$ for the intercept. An active main effect therefore has variance $c\,\sigma^2$ and an active interaction $c\,\rho\,\sigma^2$, where $\rho$ is the fixed variance ratio of an interaction to a main effect. The intercept has variance $\tau_0^2\sigma^2$, with $\tau_0^2$ fixed. Because every slab variance is proportional to $\sigma^2$, the slab scale $c$ measures effect size in units of the noise variance. The remaining symbols are fixed hyperparameters, listed with the formal statement of the prior in Appendix~\ref{app:prior} (Equation~\ref{eq:prior}). A dataset is thus drawn in three steps: the predictor matrix $\mathbf{X}$ from the predictor prior, the active set $\gamma$ from the active-effects prior, and the slab scale $c$ and the noise variance $\sigma^2$ from the effect-size prior. Given these, $\beta$ and $\mathbf{y}$ follow Equation~\ref{eq:model}. The prior was built to match real predictor matrices in the social sciences and to be conjugate, so that for small $p$ the exact posterior is available for validation (Section~\ref{sec:eval}).

\textbf{Predictor prior.}
The number of rows $n$ is drawn uniformly between \trainNMin{} and \trainNMax{}, and, independently, the number of predictors $p$ (i.e., features) uniformly between \trainPMin{} and \trainPMax{}. Figure~\ref{fig:predictor-prior} shows how the predictor matrix $\mathbf{X}$ is constructed, which mimics individual-level data of the kind collected in the social sciences: each of the $n$ rows is one independently sampled case (e.g., person, household, city, country) and each of the $p$ columns one measured characteristic. A per-dataset share $f$ of the columns forms the factor block, and the remaining columns form the generic block. The share follows a Beta distribution with a point mass at $f=0$, so the factor block is empty in \priorFactorShareZero{} of the datasets, and over all datasets the share of factor columns has median \priorFactorSharePFiveZero{}. The factor block stands in for the multi-item scales through which the social sciences measure latent constructs (e.g., IQ, social capital) \citep{borsboom2003theoretical}. The generic block represents columns that are treated as manifest variables but still correlate with one another. Its columns are jointly Gaussian with a random correlation matrix \citep{davies2000numerically}, and each also receives a small loading on the factors. The column order is then shuffled, so position carries no information. A Gaussian copula \citep{cario1997modeling} maps each column to a marginal drawn from Gaussian, Likert, binary, count and skewed families. These distributions are the typical scales found in experimental and observational social-science data \citep{likert1932technique,micceri1989unicorn,Coxe17022009,cain2017univariate}. Finally, every column is z-standardized. Appendix~\ref{app:predictor-prior} states every step in further detail.

\textbf{Active-effects prior.}
Every dataset follows strong heredity: an interaction is eligible only when both of its parent main effects, those of its two predictors, are active. The active-effects prior mixes two ways of drawing $\gamma$ (Figure~\ref{fig:active-outcome}). Under the \emph{count prior}, the number $k_{\mathrm{main}}$ of active main effects is drawn from a zero-inflated truncated geometric distribution, whose point mass at zero is the share of datasets with no active effect, and $k_{\mathrm{main}}$ main effects are picked uniformly among the $p$ candidates. The number of active interactions is drawn from the same family, and that many pairs are picked uniformly among the eligible ones. The number of active effects therefore does not grow with the number of predictors \citep[see][]{ley2009effect}. Under the \emph{rate prior}, one inclusion rate for main effects and one for interactions are drawn per dataset, and every candidate main effect, then every eligible pair, is included with that probability, so more candidates yield more active effects. Each dataset follows the count prior with probability \hpCountShare{} and the rate prior otherwise, so $\pi(\gamma \mid p)$ is the average of the two priors (Appendix~\ref{app:active-set}). The share of datasets with no active main effect is \priorNullFrac{}.

\textbf{Effect-size prior.}
Drawing $c$ per dataset rather than fixing it follows the mixtures of $g$-priors \citep{liang2008mixtures}. The hyperprior of $c$ is calibrated so that the standardized effect of an active main effect, $|\beta_j|\,\mathrm{sd}(\mathbf{x}_j)/\mathrm{sd}(\mathbf{y})$, matches meta-analytic norms for published correlations in psychology: prior quartiles of \priorMainsPTwoFive{}, \priorMainsPFiveZero{} and \priorMainsPSevenFive{} against norms of \priorNormPTwoFive{}, \priorNormPFiveZero{} and \priorNormPSevenFive{} \citep{gignac2016effect}. The ratio $\rho$ is fixed at \hpRho{}, so that interaction effects are smaller than main effects.


\subsection{Network and training}
\label{sec:model}

\begin{figure}[t]
\begin{center}
\includegraphics[width=\linewidth]{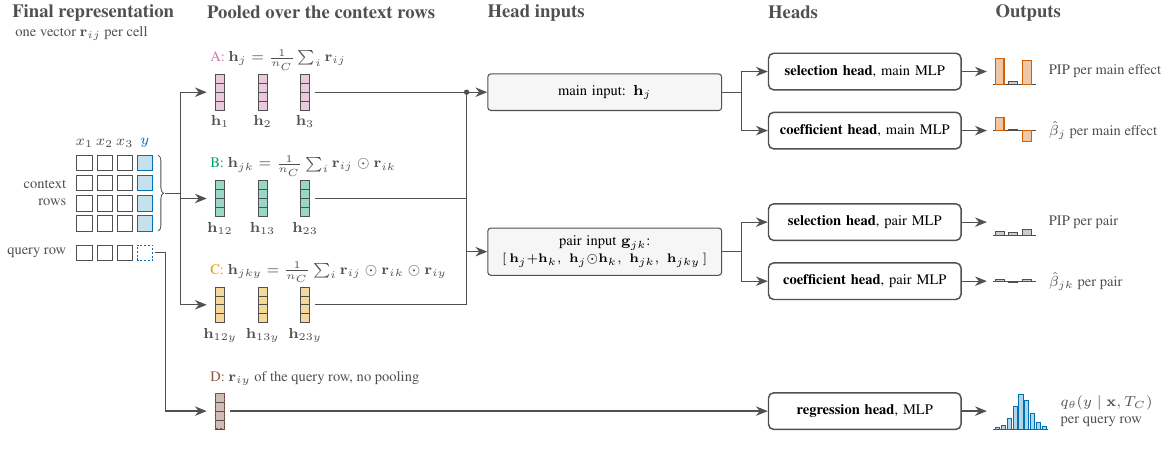}
\end{center}
\caption{The three read-out heads. Left: the final representation, one vector $\mathbf{r}_{ij}$ per cell in row $i$ and column $j$. Middle: the pooled vectors of Equation~\ref{eq:pooled}. Path A averages each predictor column over the context rows into $\mathbf{h}_j$. Paths B and C are the \emph{combine-then-mean summaries}, which multiply the cells of two predictor columns within each row, in C also with the outcome cell, and then average into $\mathbf{h}_{jk}$ and $\mathbf{h}_{jky}$. Path D is the outcome cell $\mathbf{r}_{iy}$ of a query row, not pooled. Right: the selection and coefficient heads each use a main MLP on $\mathbf{h}_j$ and a pair MLP on $\mathbf{g}_{jk}$ (Equation~\ref{eq:pair-input}). The regression head maps $\mathbf{r}_{iy}$ of each query row to a bar distribution.}
\label{fig:heads}
\end{figure}

\textbf{Architecture.}
\textsc{LinearPFN} is based on \textsc{nanoTabPFN} \citep{pfefferle2025nanotabpfn}. Each cell of the table is embedded into a vector of dimension $D = \modelEmb{}$, $\mathbf{u}_{ij} \in \R^{D}$ for predictor $j$ in row $i$ and $\mathbf{u}_{iy} \in \R^{D}$ for the outcome, which enters as one more column and is masked for the query rows. The backbone stacks \modelLayers{} transformer blocks, and each block updates every cell vector in three steps. First, in attention across features, each cell attends to the cells of its row. Second, in attention across rows, each cell attends to the cells of its column in the context rows. Third, an MLP transforms each cell vector separately. The output of the last block is the final representation (see Figure~\ref{fig:visual-abstract}, bottom).

\textbf{Three heads and their inputs.}
Let $\mathbf{r}_{ij}, \mathbf{r}_{iy} \in \R^{D}$ denote the cell vectors after the last block, the final representation. The selection and coefficient heads need one input vector per candidate effect, computed from the context rows $i = 1, \dots, n_C$. We pool in two orders, averaging a column directly or averaging a row-wise product of columns:
\begin{equation}
\mathbf{h}_j = \frac{1}{n_C}\sum_{i=1}^{n_C} \mathbf{r}_{ij},
\qquad
\mathbf{h}_{jk} = \frac{1}{n_C}\sum_{i=1}^{n_C} \mathbf{r}_{ij}\odot\mathbf{r}_{ik},
\qquad
\mathbf{h}_{jky} = \frac{1}{n_C}\sum_{i=1}^{n_C} \mathbf{r}_{ij}\odot\mathbf{r}_{ik}\odot\mathbf{r}_{iy},
\label{eq:pooled}
\end{equation}
where $\odot$ is the elementwise product. A main effect has its own column, and its input is $\mathbf{h}_j$. An interaction has no column of its own. The table holds $x_j$ and $x_k$ but not their product, so no cell belongs to the interaction $x_j x_k$. Its input is assembled from the columns $j$ and $k$ in four parts,
\begin{equation}
\mathbf{g}_{jk} = \big[\,\mathbf{h}_j + \mathbf{h}_k,\ \ \mathbf{h}_j \odot \mathbf{h}_k,\ \ \mathbf{h}_{jk},\ \ \mathbf{h}_{jky}\,\big],
\label{eq:pair-input}
\end{equation}
where $[\cdot]$ concatenates, so that $\mathbf{h}_j, \mathbf{h}_{jk}, \mathbf{h}_{jky} \in \R^{D}$ and $\mathbf{g}_{jk} \in \R^{4D}$. Every part is symmetric in $j$ and $k$, so a pair receives the same input in either order. The first two parts are \emph{mean-then-combine} summaries ($\mathbf{h}_j + \mathbf{h}_k,\ \ \mathbf{h}_j \odot \mathbf{h}_k$): each column is averaged over the rows before the two means are combined. They carry what the backbone has stored about each parent predictor on its own, which the interaction needs because under strong heredity it can be active only if both parents are. They cannot carry the evidence for the interaction itself, because in general $\mathbf{h}_j \odot \mathbf{h}_k \neq \mathbf{h}_{jk}$: the product of two means differs from the mean of the products. The last two parts are \emph{combine-then-mean} summaries ($\mathbf{h}_{jk},\ \ \mathbf{h}_{jky}$), which multiply within each row and average afterwards. 

Three heads read the final representation of the shared backbone in the same forward pass (see Figure~\ref{fig:heads}, Appendix~\ref{app:model}). Each targets a different posterior quantity:

\begin{enumerate}
\item \emph{Regression head.} It maps the outcome cell $\mathbf{r}_{iy}$ of each query row $i$ to a bar distribution, as in \textsc{TabPFN} \citep{pfefferle2025nanotabpfn,hollmann2022tabpfn}: the range of possible outcomes is divided into fixed bins and the head predicts the probability of each bin, which gives a piecewise-constant density over $y$. Its loss is minimized in expectation by the posterior predictive distribution under the prior \citep{müller2024transformersbayesianinference}. The other two heads cannot replace it, because they supply neither the noise variance nor the posterior covariance of the coefficients (Appendix~\ref{app:predictive}). Removing its loss also lowers selection and coefficient accuracy (Appendix~\ref{app:two-head}).

\item \emph{Selection head.} This head returns a \ac{pip} for every candidate effect. One MLP scores each main effect from $\mathbf{h}_j$, and a second MLP scores each interaction from $\mathbf{g}_{jk}$. A \ac{pip} is the posterior probability that an effect is active. 

\item \emph{Coefficient head.} Like the selection head, it takes $\mathbf{h}_j$ and $\mathbf{g}_{jk}$ as its inputs for the two MLPs, and returns one number per effect, the posterior mean $\E[\beta_e \mid T]$. This is the marginal mean, averaged over all models. The conditional mean $\E[\beta_e \mid T, \gamma_e = 1]$ is instead the size of the effect given that it is active. The two are linked by the \ac{pip}, $\E[\beta_e \mid T] = \Pr(\gamma_e = 1 \mid T)\,\E[\beta_e \mid T, \gamma_e = 1]$, so an effect with a \ac{pip} of one half receives half its conditional mean. The output is dense, so every candidate effect receives a value and none is set to zero by a selection step.

\end{enumerate}

\textbf{Loss.}
Training minimizes, over datasets drawn from the prior, the sum of three terms:
\begin{equation}
\mathcal{L} = \underbrace{-\tfrac{1}{|Q|}\textstyle\sum_{i\in Q}\log q_\theta(y_i \mid \mathbf{x}_i, T_C)}_{\text{regression head:  NLL}}
+ \underbrace{\tfrac{1}{d-1}\textstyle\sum_{e}\text{\acs{bce}}(\mathrm{sigmoid}(s_e),\gamma_e)}_{\text{selection head: BCE}}
+ \underbrace{\tfrac{1}{d-1}\textstyle\sum_{e}(b_e - \beta_e)^2}_{\text{coefficient head: MSE}} .
\label{eq:loss}
\end{equation}
Here $T_C$ is the context part of a dataset, $Q$ its query rows, $(\mathbf{x}_i, y_i)$ the predictors and outcome of query row $i$, and $q_\theta$ the bar distribution of the regression head with network weights $\theta$. The sums over $e$ run over the $d-1$ effects other than the intercept. For effect $e$, $s_e$ is the selection logit, $b_e$ the coefficient output, and $\gamma_e$ and $\beta_e$ the active indicator and coefficient of Equation~\ref{eq:model}, known because the dataset was simulated. The first term is the loss of a \ac{pfn}, whose expected minimizer is the posterior predictive distribution under the prior \citep{müller2024transformersbayesianinference}. The second is a proper scoring rule whose expected minimizer is the posterior probability that the effect is active, so the head is trained toward the exact \acp{pip}, and that of the third is the posterior mean of the coefficient.

\textbf{Training and prediction.}
\textsc{LinearPFN} has \modelParams{} parameters and was trained for \trainSteps{} optimizer steps at an effective batch of \trainEffBatch{} datasets, \trainDatasetsSeenBig{} datasets in total, on \trainGpuCountWord{} \trainGpu{} GPUs (model card, optimizer and schedule in Appendix~\ref{app:model}, Table~\ref{tab:model-card}). Fitting a dataset is one forward pass that returns the \acp{pip} and the coefficients. Following the \ac{mpm} \citep{barbieri2004optimal}, \textsc{LinearPFN} selects every effect whose \ac{pip} exceeds one half.


\subsection{Evaluation}
\label{sec:eval}

\textbf{Exact posterior.}
Given the active set $\gamma$ and the slab scale $c$, the prior is conjugate, so the marginal likelihood of every model has a closed form (Appendix~\ref{app:exact}). For $p \leq \exactMaxP{}$ there are at most \exactNModelsMaxP{} models, few enough to sum over all of them and to integrate out $c$ numerically. This yields the exact posterior, the probability of every model, and from it the exact \acp{pip} and posterior-mean coefficients. It uses the same implementation of the prior as the synthetic training data. We compare \textsc{LinearPFN} with it on held-out prior draws, on six criteria. They cover the predictive distribution, the coefficients and the \acp{pip} (Table~\ref{tab:gates}, Appendix~\ref{app:validation}).

\textbf{Benchmarks.}
Practical performance is measured on three benchmarks using real predictor matrices:

\emph{Real predictor matrices, outcomes from the prior} is the main benchmark. The predictor matrices are the \rxPoolN{} tables of the social-science packages in the \textsc{Rdatasets} collection \citep{Rdatasets,Rlang} that pass fixed inclusion and cleaning rules. A table with more than \rxPoolNMax{} rows or \rxPoolPMax{} predictors is reduced to a random subset of that size, drawn once with a fixed seed (Appendix~\ref{app:predictor-matrices}). Each dataset draws a predictor matrix uniformly from the \rxPoolN{} predictor matrices and an outcome from the prior of Section~\ref{sec:prior} on the standardized columns of that predictor matrix, $\mathbf{y} = \mathbf{Z}\beta + \epsilon$, so the active set and the coefficients are known and every score is computed against them. Each draw has a density, the share $k_{\mathrm{main}}/p$ of predictors with an active main effect, and an $R^2 = \mathrm{var}(\mathbf{Z}\beta)/\mathrm{var}(\mathbf{y})$, the share of outcome variance explained by $\mathbf{Z}\beta$ in the drawn dataset. High density and high $R^2$ are rare under the prior, so unstratified draws would not represent those regimes with enough datasets to compare methods within them. Each draw is therefore assigned to its cell in a $\ipLevelsRealX{} \times \ipLevelsRealX{}$ grid of density by $R^2$ bins, and draws continue until every cell holds \ipPerCellRealX{} datasets, \ipDatasetsOursRealX{} in all, so that sparse and dense active sets and weak and strong signal are represented equally (Appendix~\ref{app:panel-construction}). Every method receives the same standardized predictors and the same candidate set of $p + \binom{p}{2}$ effects. The baselines are the lasso \citep{tibshirani1996regression}, stability selection \citep{meinshausenStabilitySelection2010,shah2013variable}, \textsc{SuSiE} \citep{susieR1,susieR2}, \textsc{glinternet} \citep{gli2026} and \textsc{hierNet} \citep{hier2020}, each with its own selection rule and ranking score. \textsc{LinearPFN} selects by the \ac{mpm} rule and ranks by \ac{pip} (Appendix~\ref{app:method-set}). Selection is scored per dataset by the $F_1$ of each method's own hard selection against the true active set and by the \ac{auc} with which its ranking separates active from inactive candidate effects. Coefficients are scored by the \ac{rmse} against the true coefficient vector over every candidate effect. Every metric is averaged over datasets (Appendix~\ref{app:metrics}).

\begin{table}[t]
\caption{Validation of \textsc{LinearPFN} against the exact posterior on held-out prior draws with $p \le \exactMaxP{}$ (coverage on every $p$). The thresholds were fixed before the paper's model was trained.}
\label{tab:gates}
\centering
\small

\begin{tabular*}{\textwidth}{@{}l@{\extracolsep{\fill}}rrlrr@{}}
\toprule
\multicolumn{3}{@{}l}{\emph{predictive distribution and coefficients}} & \multicolumn{3}{l@{}}{\emph{inclusion probabilities}} \\
\cmidrule(r){1-3}\cmidrule(l){4-6}
criterion & threshold & value & criterion & threshold & value \\
\midrule
headroom & $\geq$ 0.95 & 0.9605 & AUC gap & $\geq -0.01$ & \ensuremath{-0.0026} \\
coverage error & $\leq$ 0.03 & 0.0005 & ECE gap & $\leq$ 0.01 & 0.0034 \\
coefficient correlation & $\geq$ 0.99 & 0.9931 & resolution ratio & $\geq$ 0.90 / 0.85 & 0.9640 \\
\bottomrule
\end{tabular*}

\end{table}
\emph{Real predictor matrices, outcomes outside the prior} is an ablation of the outcome model. It takes the same \ipDatasetsOursRealX{} datasets as the main benchmark and replaces one stage of the outcome model of Equation~\ref{eq:model} at a time, holding every other quantity of the draw fixed. Given the draw $(\mathbf{X}, \gamma, c, \sigma^2, \beta)$ of a main-benchmark dataset, each variant keeps the predictor matrix $\mathbf{X}$, the active main effects, the slab scale $c$ and the noise variance $\sigma^2$, and redraws the outcome as $\mathbf{y}' = \mathbf{Z}\beta' + \epsilon'$ with fresh noise from a random stream of its own. We test three variants.

\begin{enumerate}
\item \textbf{Fixed-magnitude coefficients.} In the main benchmark each active coefficient is drawn from a normal distribution, $\beta_e \sim \gN(0,\, v_e\,\sigma^2)$. Here every active effect other than the intercept is set to $\beta'_e = s_e\sqrt{v_e\,\sigma^2}$ with a random sign $s_e \in \{-1, +1\}$. 
\item \textbf{Many large interactions.} In the main benchmark interactions are rare. The median dataset has \ipIntCountMedianRealX{} active interaction (mean \ipIntCountMeanRealX{}), \ipIntNoneShareRealX{} of the datasets have none, and each interaction has half the slab variance of a main effect ($\rho = \hpRho{}$). This variant changes both properties. First, every pair whose two parent main effects are active becomes an interaction independently with probability one half, $\gamma'_{jk} \sim \mathrm{Bernoulli}(1/2)$, so the count grows with the number of active main effects. Second, the slab variance of an interaction is raised from $c\rho$ to $c$, that is $\rho = 1$, so an interaction is on average as large as a main effect. The main-effect coefficients are kept, so strong heredity still holds.
\item \textbf{Heavy-tailed noise.} In the main benchmark the noise of row $i$ is $\epsilon_i = \sigma z_i$ with $z_i \sim \gN(0, 1)$. Here $z_i$ is replaced by $t_i/\sqrt{3}$, where $t_i$ is drawn from a Student-$t$ distribution with three degrees of freedom. That distribution has variance 3, so $t_i/\sqrt{3}$ has unit variance like $z_i$, and the noise variance stays $\sigma^2$, but a few rows receive much larger errors, which act as outliers. The active set and the coefficients are unchanged.
\end{enumerate}

\emph{Real outcomes} scores \textsc{LinearPFN} where no active set is known. We take \realyN{} published tables that are widely used to demonstrate regression methods (Table~\ref{tab:real-y}). Here the reference is an \ac{mcmc} sampler over the model space \citep{madigan1995bayesian} that targets the same posterior. Agreement is measured by the correlation and \ac{rmse} between the \acp{pip} of \textsc{LinearPFN} and of the sampler and between their posterior-mean coefficients over every candidate effect (Appendix~\ref{app:real-y}).

\begin{table}[t]
\caption{Benchmark on the real predictor matrices, \ipSupportNRealX{} datasets with outcomes drawn from the prior. Over datasets, AUC and $F_1$ computed per dataset (/ds), with the first and third quartiles in brackets. Fit seconds are one fit on one CPU thread. On one \fitGpuName{} GPU, the same forward pass of \textsc{LinearPFN} takes \fitSecGpu{} seconds at the median [\fitSecGpuQOne{}, \fitSecGpuQThree{}].}
\label{tab:ip-summary}
\small
\begin{tabular*}{\textwidth}{l@{\extracolsep{\fill}}rrrr}
\toprule
method & AUC/ds, mean & $F_1$/ds, mean & coef RMSE, mean & fit s, median \\
\midrule
LinearPFN & \textbf{\ipAucMeanOursRealX{}} \iqr{\ipAucQOneOursRealX{}}{\ipAucQThreeOursRealX{}} & \textbf{\ipFOneMeanOursRealX{}} \iqr{\ipFOneQOneOursRealX{}}{\ipFOneQThreeOursRealX{}} & \ipRmseMeanOursRealX{} \iqr{\ipRmseQOneOursRealX{}}{\ipRmseQThreeOursRealX{}} & \ipFitMedOursRealX{} \iqr{\ipFitQOneOursRealX{}}{\ipFitQThreeOursRealX{}} \\
hierNet & \ipAucMeanHierNetRealX{} \iqr{\ipAucQOneHierNetRealX{}}{\ipAucQThreeHierNetRealX{}} & \ipFOneMeanHierNetRealX{} \iqr{\ipFOneQOneHierNetRealX{}}{\ipFOneQThreeHierNetRealX{}} & \ipRmseMeanHierNetRealX{} \iqr{\ipRmseQOneHierNetRealX{}}{\ipRmseQThreeHierNetRealX{}} & \ipFitMedHierNetRealX{} \iqr{\ipFitQOneHierNetRealX{}}{\ipFitQThreeHierNetRealX{}} \\
glinternet & \ipAucMeanGlinternetRealX{} \iqr{\ipAucQOneGlinternetRealX{}}{\ipAucQThreeGlinternetRealX{}} & \ipFOneMeanGlinternetRealX{} \iqr{\ipFOneQOneGlinternetRealX{}}{\ipFOneQThreeGlinternetRealX{}} & \ipRmseMeanGlinternetRealX{} \iqr{\ipRmseQOneGlinternetRealX{}}{\ipRmseQThreeGlinternetRealX{}} & \ipFitMedGlinternetRealX{} \iqr{\ipFitQOneGlinternetRealX{}}{\ipFitQThreeGlinternetRealX{}} \\
SuSiE & \ipAucMeanSusieRealX{} \iqr{\ipAucQOneSusieRealX{}}{\ipAucQThreeSusieRealX{}} & \ipFOneMeanSusieRealX{} \iqr{\ipFOneQOneSusieRealX{}}{\ipFOneQThreeSusieRealX{}} & \ipRmseMeanSusieRealX{} \iqr{\ipRmseQOneSusieRealX{}}{\ipRmseQThreeSusieRealX{}} & \ipFitMedSusieRealX{} \iqr{\ipFitQOneSusieRealX{}}{\ipFitQThreeSusieRealX{}} \\
lasso & \ipAucMeanLassoRealX{} \iqr{\ipAucQOneLassoRealX{}}{\ipAucQThreeLassoRealX{}} & \ipFOneMeanLassoRealX{} \iqr{\ipFOneQOneLassoRealX{}}{\ipFOneQThreeLassoRealX{}} & \ipRmseMeanLassoRealX{} \iqr{\ipRmseQOneLassoRealX{}}{\ipRmseQThreeLassoRealX{}} & \textbf{\ipFitMedLassoRealX{}} \iqr{\ipFitQOneLassoRealX{}}{\ipFitQThreeLassoRealX{}} \\
stability & \ipAucMeanStabilityRefitRealX{} \iqr{\ipAucQOneStabilityRefitRealX{}}{\ipAucQThreeStabilityRefitRealX{}} & \ipFOneMeanStabilityRefitRealX{} \iqr{\ipFOneQOneStabilityRefitRealX{}}{\ipFOneQThreeStabilityRefitRealX{}} & \ipRmseMeanStabilityRefitRealX{} \iqr{\ipRmseQOneStabilityRefitRealX{}}{\ipRmseQThreeStabilityRefitRealX{}} & \ipFitMedStabilityRefitRealX{} \iqr{\ipFitQOneStabilityRefitRealX{}}{\ipFitQThreeStabilityRefitRealX{}} \\
\bottomrule
\end{tabular*}
\end{table}


\section{Results}
\label{sec:results}

\textbf{Fidelity to the exact posterior.}
\textsc{LinearPFN} meets all six validation criteria (Table~\ref{tab:gates}). Its gap to the exact posterior is \gateAucGap{} in the \ac{auc} of the \acp{pip}, and its posterior-mean coefficients correlate with the exact ones at \gateCoefRExactPanel{}. On the \ipDatasetsExactRealX{} benchmark datasets within enumeration reach, its $F_1$ differs from that of the exact posterior by \ipDiffFOneExactRealX{} and its coefficient \ac{rmse} by \ipDiffRmseExactRealX{} (Appendix~\ref{app:exact-fidelity}). Beyond that reach, on \ipMcmcNRealX{} main-benchmark datasets, its $F_1$ differs from that of the \ac{mcmc} sampler by \ipDiffFOneMcThreeRealX{} [\ipDiffLoFOneMcThreeRealX{}, \ipDiffHiFOneMcThreeRealX{}] and its \ac{auc} by \ipDiffAucMcThreeRealX{}.

\textbf{Real predictor matrices, outcomes from the prior.}
Table~\ref{tab:ip-summary} reports every method on the \ipDatasetsOursRealX{} datasets over the real predictor matrices (for paired differences see Appendix~\ref{app:main-bench}). Figure~\ref{fig:ip-sweeps} reports $F_1$, \ac{auc} and \ac{rmse} as a function of the number of predictors, number of rows, density and $R^2$.

\emph{Selection is ahead of every baseline in every slice.} The $F_1$ margin grows with $p$ and is widest at \ipMarginFOneNMaxBinRealX{} rows, and the \ac{auc} margin shrinks as $n$ grows (Figure~\ref{fig:ip-sweeps}). The lasso, \textsc{glinternet} and \textsc{hierNet} select many inactive effects, and \textsc{SuSiE} and stability selection miss more active ones. At high density \textsc{hierNet} and \textsc{glinternet} catch up. Against \textsc{SuSiE}, the closest baseline, the paired difference in $F_1$ per dataset is \ipDiffFOneSusieRealX{} [\ipDiffLoFOneSusieRealX{}, \ipDiffHiFOneSusieRealX{}] (bootstrap 95\,\% interval).

\emph{Coefficient error matches the best baseline.} Except for \textsc{SuSiE}, the baselines return coefficients that are exactly zero outside their selected set, while \textsc{LinearPFN} returns the dense posterior mean. We also score \textsc{LinearPFN} hard-zeroed, setting to zero every coefficient whose \ac{pip} does not exceed 0.5. Against \textsc{hierNet}, the best baseline, the paired difference in \ac{rmse} is \ipDiffRmseHierNetRealX{} [\ipDiffLoRmseHierNetRealX{}, \ipDiffHiRmseHierNetRealX{}] for the dense posterior mean and \ipHzDiffRmseHierNetRealX{} [\ipHzDiffLoRmseHierNetRealX{}, \ipHzDiffHiRmseHierNetRealX{}] hard-zeroed. For $p \le \exactMaxP{}$, the exact posterior has a coefficient \ac{rmse} of \ipRmseMeanExactRealX{}, and \textsc{LinearPFN} exceeds it by \ipDiffRmseExactRealX{} (Appendix~\ref{app:exact-fidelity}).

\emph{\textsc{LinearPFN} enables GPU acceleration.} On one CPU thread \textsc{LinearPFN} takes \ipFitMedOursRealX{} seconds at the median and is slower than every baseline except \textsc{hierNet} (Table~\ref{tab:ip-summary}). Its forward pass also runs on a GPU, which the reference implementations of the baselines do not support, and on one \fitGpuName{} it takes \fitSecGpu{} seconds. The baselines ran on one CPU thread, so this comparison includes the hardware. \ac{mcmc} takes \ipFitMedMcThreeRealX{} seconds on one CPU thread, \ipFitRatioMcThreeRealX{} times as long as \textsc{LinearPFN} on the same thread and \fitMcmcOverGpu{} times its GPU forward pass, on the same datasets.

\begin{figure}[t]
\begin{center}
\includegraphics[width=\linewidth]{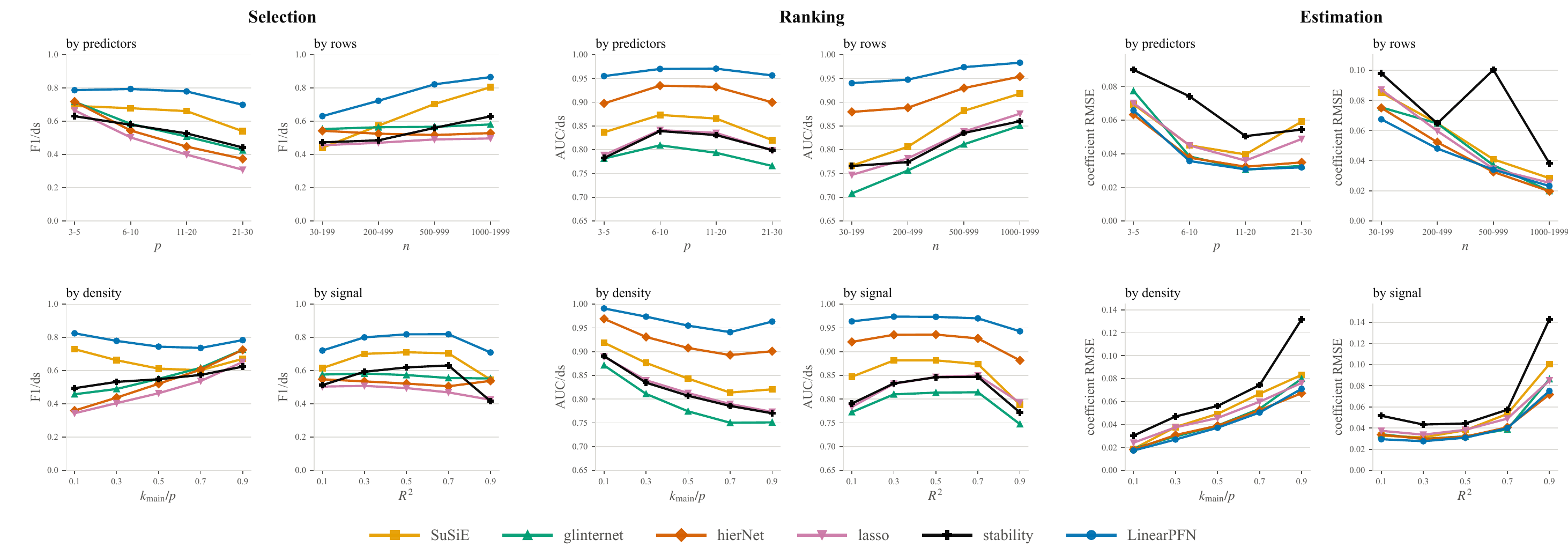}
\end{center}
\caption{Benchmark on the \rxPoolN{} real predictor matrices, \ipDatasetsOursRealX{} datasets with outcomes drawn from the prior. Left: selection, $F_1$ per dataset at each method's own selection rule. Middle: ranking, AUC per dataset. Right: estimation, coefficient RMSE over every candidate effect. Each panel slices the same datasets by the number of predictors, the number of rows, the density $k_{\mathrm{main}}/p$ and the $R^2$; every point is the mean over the datasets of that slice. Density and $R^2$ are binned in the \ipLevelsRealX{} bins of the benchmark grid, each \ipBinWidthRealX{} wide, and the axis marks the lower edge of each bin (Appendix~\ref{app:panel-construction}).}

\label{fig:ip-sweeps}
\end{figure}

\textbf{Real predictor matrices, outcomes outside the prior.}
We test three variants of the outcome model (Equation~\ref{eq:model}): fixed-magnitude coefficients, many large interactions and heavy-tailed noise (Appendix~\ref{app:knobs}). Table~\ref{tab:knobs} reports every method on each variant compared with the control.

\emph{\textsc{LinearPFN} keeps the highest \ac{auc} and $F_1$ in every variant.} The smallest $F_1$ margin is over \textsc{SuSiE} under many large interactions, \knobDiffFOneSusieInt{} [\knobDiffLoFOneSusieInt{}, \knobDiffHiFOneSusieInt{}]. The one metric on which \textsc{LinearPFN} is not first is coefficient \ac{rmse} under many large interactions. There its error of \knobRmseOursInt{} is above that of three baselines, and the largest gap is to \textsc{glinternet}, \knobDiffRmseGlinternetInt{} [\knobDiffLoRmseGlinternetInt{}, \knobDiffHiRmseGlinternetInt{}].

\emph{Many large interactions cost coefficient accuracy.} They raise the coefficient error of every method, that of \textsc{LinearPFN} by \knobCtrlDiffRmseOursInt{} [\knobCtrlDiffLoRmseOursInt{}, \knobCtrlDiffHiRmseOursInt{}], and lower the \ac{auc} of every method. Heavy-tailed noise changes the coefficient error of \textsc{LinearPFN} by \knobCtrlDiffRmseOursNoise{} [\knobCtrlDiffLoRmseOursNoise{}, \knobCtrlDiffHiRmseOursNoise{}] and raises $F_1$ slightly for every method. Only stability selection loses coefficient accuracy under it, through a few diverging refits that raise its mean error by \knobCtrlDiffRmseStabilityRefitNoise{} (Table~\ref{tab:knobs}).

\textbf{Real outcomes.}
On \realyN{} published tables with their actual outcome column, we compare \textsc{LinearPFN} with the \ac{mcmc} sampler that targets the same posterior (Table~\ref{tab:real-y}, Appendix~\ref{app:real-y}). The \acp{pip} of the two correlate at \realyPipRAll{} pooled over the \realyMpmSlots{} candidate effects, and the two \acp{mpm} are identical on \realyMpmIdentical{} of the \realyN{} tables. These tables have no known active set, so they do not measure selection accuracy.


\section{Discussion}
\label{sec:discussion}
We identify four limitations of our amortized variable-selection method. First, when two predictors are nearly collinear, the posterior mass splits across them, both \acp{pip} can fall below one half, and the \ac{mpm} rule then selects neither \citep{susieR2}. Unlike \textsc{SuSiE} \citep{susieR1}, \textsc{LinearPFN} reports no credible sets that group such predictors. Second, without a GPU, \textsc{LinearPFN} is slow. On one CPU thread it is the second slowest of the six methods, ahead of \textsc{hierNet} only. However, neither the cost on a CPU nor the fixed prior is intrinsic to the approach. Pruning, quantization and knowledge distillation are established routes to a cheaper forward pass, which we did not pursue. Third, in coefficient error \textsc{LinearPFN} only matches the best baseline, although it leads in selection. Fourth, the network is trained on a single prior, which assumes strong heredity, Gaussian noise, numeric predictors and independent rows. Its lead in selection held when the coefficients, the interactions or the noise departed from this prior. Follow-up work could condition the network on the prior's hyperparameters, so that users can state their own prior without retraining.

\textsc{LinearPFN} is a transformer pretrained on an explicit conjugate spike-and-slab prior over main effects and pairwise interactions. Its outputs are validated against the exact posterior wherever that posterior can be enumerated, and they are compared with an \ac{mcmc} sampler on real outcomes. On real predictor matrices with outcomes drawn from its prior it selects better than every baseline. Amortization reduces variable selection to one forward pass, which returns the selected variables and their coefficients in \fitSecGpu{} seconds at the median on one GPU. The \ac{mcmc} sampler over the same posterior needs \ipFitMedMcThreeRealX{} seconds at the median on one CPU thread and meets its convergence criteria on \ipMcmcConvPctRealX{} of runs. Bayesian model averaging with interactions thereby becomes feasible where it has to be repeated many times.

\newpage

\subsection*{AI use statement}
The authors developed the theoretical models, conceptual frameworks, and research methodology. We used generative AI tools to discuss and give feedback on these models and on the design of the experiments, to implement methods, and to clean and reformat datasets. We used AI tools to find related work. The authors read the cited papers and checked each citation. We used AI tools to correct grammar, spelling, and wording. For interpretation, we used AI tools to suggest possible explanations for observed patterns. The authors checked these suggestions against the data and decided which conclusions to report. We did not use generative AI tools to provide critical steps in proofs of mathematical claims, to formulate mathematical claims, to generate synthetic datasets, to translate text, to support qualitative or thematic data analysis, or to propose or refine hypotheses. The authors reviewed all AI-assisted work, including all AI-generated code. We take responsibility for the final content of this work, including text, claims, and artifacts produced with the aid of generative AI. The synthetic datasets are drawn by code from the prior stated in Section~\ref{sec:prior} and were not produced by a generative model.

\subsection*{Ethics statement}
This work uses publicly available data only: \rxPoolN{} tables of the \textsc{Rdatasets} collection as predictor matrices, and \realyN{} published regression datasets with their actual outcome (Appendix~\ref{app:real-y}). No data were collected from human subjects for this work, and no personal data are processed beyond what the sources already publish. The method is a statistical tool for variable selection. Its outputs are probabilities under a stated prior and do not by themselves support causal claims.

\subsection*{Reproducibility statement}

The prior is specified in Section~\ref{sec:prior} and Appendix~\ref{app:prior}, with every fixed value in Table~\ref{tab:prior-config}. The exact posterior and the MCMC sampler are described in Appendix~\ref{app:exact}, the network, its training and the model card in Appendix~\ref{app:model} and Table~\ref{tab:model-card}, and the validation criteria in Appendix~\ref{app:validation}. The construction of every benchmark, the methods and the metrics are given in Appendices~\ref{app:benchmark} and~\ref{app:bench-results}. The predictor matrices come from a public collection and are selected and cleaned by fixed rules (Appendix~\ref{app:predictor-matrices}), and the real-outcome tables are listed with their preparation in Appendix~\ref{app:real-y}. The network is implemented in PyTorch \citep{paszke2019pytorchimperativestylehighperformance}, and the exact posterior and the MCMC sampler in NumPy \citep{harris2020array} and SciPy \citep{2020SciPy-NMeth}. The lasso and stability selection use scikit-learn \citep{pedregosa2011scikit}, and the other baselines run in R \citep{Rlang}. The code, with every experiment and the scripts that produce each number, table and figure, is available at \url{https://github.com/schiekiera/LinearPFN}, and the trained model at \url{https://huggingface.co/schiekiera/LinearPFN}.

\subsubsection*{Acknowledgments}
This research was partially supported by the Deutsche Forschungsgemeinschaft (DFG) through the projects ``What's in a name? Computational modeling and experimental investigations on the non-arbitrariness of word label choices'' (project number 459717703), ``A computational implementation of the Swinging Lexical Network model of language production'' (project number 532390335), the DFG Cluster of Excellence MATH+ (EXC-2046/2, project id 390685689), as well as by the German Federal Ministry of Research, Technology and Space (research campus Modal, fund number 05M14ZAM, 05M20ZBM) and the VDI/VDE Innovation + Technik GmbH (fund number 16IS23025B). We would also like to thank the Speinshart Scientific Center for AI and SuperTech, where part of this work was conducted, for providing such a stimulating research environment.


\bibliography{references}
\bibliographystyle{iclr2027_conference}

\newpage



\appendix


\section{Glossary}
\label{app:glossary}

\subsection{Abbreviations}
\label{app:abbreviations}
\printacronyms[heading=none]

\subsection{Notation}
\label{app:notation}

\textbf{Datasets and predictors.}
\begin{itemize}
\item $T$: a dataset, a table of $n$ rows. $T_C$: its context rows. $Q$: the index set of its query rows.
\item $n$, $p$: the number of rows and of predictors of a dataset. $n_C$: the number of context rows.
\item $i$: the row index, $i = 1, \dots, n$. $j$, $k$: predictor indices.
\item $\mathbf{X} \in \R^{n\times p}$: the predictor matrix, with z-standardized columns $\mathbf{x}_j$. $\mathbf{x}_i$ in Equation~\ref{eq:loss}: the predictors of query row $i$.
\item $\ell$: the index of a real predictor matrix $\mathbf{X}_\ell \in \R^{n_\ell\times p_\ell}$, with expanded predictor matrix $\mathbf{Z}_\ell$.
\item $\mathbf{y}$: the outcome vector. The prior does not standardize it.
\item $d = 1 + p + \binom{p}{2}$: the number of candidate effects, intercept included. $e$: the index of a candidate effect.
\item $\mathbf{Z}$: the expanded predictor matrix, a column of ones, the columns of $\mathbf{X}$ and their pairwise products $\mathbf{x}_j\odot\mathbf{x}_k$. $\odot$: the elementwise product.
\end{itemize}

\textbf{Prior.}
\begin{itemize}
\item $\gamma \in \{0,1\}^d$: the active set, with the intercept always active. One value of $\gamma$ is a model. $\mathbf{Z}_\gamma$: the columns of $\mathbf{Z}$ of the active effects.
\item $\beta \in \R^d$: the coefficient vector, zero outside the active set. $\beta_\gamma$: its active entries.
\item $\epsilon$: the noise vector. $\sigma^2$: the noise variance, with prior $\mathrm{InvGamma}(a_0, b_0)$.
\item $\delta_0$: the point mass at zero, the spike.
\item $c$: the slab scale, the variance of an active main effect relative to $\sigma^2$, with prior $\mathrm{LogNormal}(\mu_c, s_c^2)$ truncated to $[c_{\min}, c_{\max}]$.
\item $\rho$: the fixed variance ratio of an active interaction to an active main effect.
\item $\tau_0^2$: the fixed slab variance of the intercept relative to $\sigma^2$.
\item $v_e$: the slab variance of effect $e$ relative to $\sigma^2$: $c$ for a main effect, $c\,\rho$ for an interaction and $\tau_0^2$ for the intercept.
\item $\pi(\mathbf{X} \mid n, p)$: the predictor prior.
\item $\pi(\gamma \mid p)$: the active-effects prior. $\pi_{\mathrm{count}}$, $\pi_{\mathrm{rate}}$: the count prior and the rate prior. $\alpha$: the probability of the count prior.
\item $k_{\mathrm{main}}$, $k_{\mathrm{int}}$: the numbers of active main effects and of active interactions. $M_{\mathrm{int}}$: the number of eligible pairs.
\item $\zeta_{\mathrm{main}}$, $\bar{k}_{\mathrm{main}}$, $\zeta_{\mathrm{int}}$, $\bar{k}_{\mathrm{int}}$: the zero masses and geometric means of the count prior. $\omega_{\mathrm{main}}$, $\omega_{\mathrm{int}}$: the inclusion rates of the rate prior.
\item $k_{\mathrm{main}}/p$: the density of a dataset.
\item $f$: the share of columns in the factor block. $R$: the correlation matrix of the generic block. $m$: the number of latent factors. $h^2_j$: the communality of column $j$.
\item $R^2$: $\mathrm{var}(\mathbf{Z}\beta)/\mathrm{var}(\mathbf{y})$ over the rows of a dataset.
\end{itemize}

\textbf{Network and training.}
\begin{itemize}
\item $\mathbf{u}_{ij}$, $\mathbf{u}_{iy}$: the embedded cell of predictor $j$ and the embedded outcome cell in row $i$.
\item $\mathbf{r}_{ij}$, $\mathbf{r}_{iy}$: the same cells after the last block, the final representation. $D$: the dimension of every cell vector.
\item $\mathbf{h}_j$, $\mathbf{h}_{jk}$, $\mathbf{h}_{jky}$: the context-row means of $\mathbf{r}_{ij}$, of $\mathbf{r}_{ij}\odot\mathbf{r}_{ik}$ and of $\mathbf{r}_{ij}\odot\mathbf{r}_{ik}\odot\mathbf{r}_{iy}$ (Equation~\ref{eq:pooled}).
\item $\mathbf{g}_{jk}$: the input of the pair MLPs for the interaction of predictors $j$ and $k$ (Equation~\ref{eq:pair-input}).
\item $\theta$: the network weights. $q_\theta$: the predictive distribution of the regression head.
\item $s_e$, $b_e$: the selection logit and the coefficient output for effect $e$. $\mathrm{sigmoid}(\cdot)$: the logistic function.
\end{itemize}

\textbf{Evaluation and exact posterior.}
\begin{itemize}
\item $F_1$: the harmonic mean of precision and recall of a hard selection against the true active set.
\item $\beta'$, $\gamma'$, $\mathbf{y}'$, $\epsilon'$: the redrawn quantities of a variant with outcomes outside the prior. $z_i$, $t_i$: the standard normal and the Student-$t$ draw of the noise of row $i$.
\item $\mathbf{z}^*$, $y^*$: the expanded predictors and the outcome of a new row.
\item $w_\gamma$: the posterior probability of model $\gamma$ under the exact posterior.
\item $V_0$, $V_n$, $m_n$, $a_n$, $b_n$: the prior slab matrix and the posterior quantities of the normal-inverse-gamma regression given a model.
\end{itemize}


\section{The prior}
\label{app:prior}

A dataset is drawn in six steps:
\begin{subequations}
\label{eq:prior}
\begin{align}
n &\sim \mathrm{U}\{n_{\min},\dots,n_{\max}\}, &
p &\sim \mathrm{U}\{p_{\min},\dots,p_{\max}\}, \label{eq:prior-size}\\
\mathbf{X} &\sim \pi(\mathbf{X} \mid n, p), &
\mathbf{Z} &= [\mathbf{1},\ \mathbf{X},\ (\mathbf{x}_j\odot\mathbf{x}_k)_{j<k}], \label{eq:prior-x}\\
\gamma &\sim \pi(\gamma \mid p), & & \label{eq:prior-gamma}\\
c &\sim \mathrm{LogNormal}(\mu_c, s_c^2), &
\sigma^2 &\sim \mathrm{InvGamma}(a_0, b_0), \label{eq:prior-scales}\\
\beta_e \mid \gamma_e, c, \sigma^2 &\sim (1-\gamma_e)\,\delta_0 + \gamma_e\,\gN(0,\, v_e\,\sigma^2), &
v_e &= \begin{cases} c & \text{main effect},\\ c\,\rho & \text{interaction},\\ \tau_0^2 & \text{intercept}, \end{cases} \label{eq:prior-beta}\\
\mathbf{y} &= \mathbf{Z}\beta + \epsilon, &
\epsilon &\sim \gN(0, \sigma^2 \mathbf{I}_n). \label{eq:prior-y}
\end{align}
\end{subequations}
Equation~\ref{eq:prior-size} draws the number of rows and of predictors. Equation~\ref{eq:prior-x} draws the predictor matrix from the predictor prior (Appendix~\ref{app:predictor-prior}) and expands it by a column of ones and all pairwise products. Equation~\ref{eq:prior-gamma} draws the active set from the active-effects prior (Appendix~\ref{app:active-set}). Equation~\ref{eq:prior-scales} draws the slab scale, truncated to $[c_{\min}, c_{\max}]$, and the noise variance. Equation~\ref{eq:prior-beta} sets every inactive coefficient to zero and draws every active one from a normal slab whose variance depends on the type of effect. Equation~\ref{eq:prior-y} adds Gaussian noise to $\mathbf{Z}\beta$ to form the outcome. Table~\ref{tab:prior-config} lists every fixed value of the paper's model. The size $(n, p)$ is drawn once per training batch, so the datasets of one batch share it (Appendix~\ref{app:training}). The prior on $\sigma^2$ fixes the scale of the outcome, so the posterior changes when $\mathbf{y}$ is rescaled. On data from outside the prior, the outcome is therefore standardized before the forward pass, as for the real outcomes (Appendix~\ref{app:real-y}).

\begin{table}[t]
\caption{The prior of the paper's model: its fixed values, and the standardized effects $|\beta_e|\,\mathrm{sd}(\mathbf{z}_e)/\mathrm{sd}(\mathbf{y})$ it implies over the active effects of \priorNDraws{} prior draws, where $\mathbf{z}_e$ is the column of effect $e$ in $\mathbf{Z}$.\vspace{0.5em}}
\label{tab:prior-config}
\centering
\footnotesize
\begin{tabular*}{\textwidth}{@{}l@{\extracolsep{\fill}}l@{}}
\toprule
\multicolumn{2}{@{}l}{\emph{Predictor prior}}\\
rows $n$, predictors $p$ & U\{\trainNMin{}, \ldots, \trainNMax{}\}, U\{\trainPMin{}, \ldots, \trainPMax{}\}\\
factor share $f$ & $0$ with probability \hpFZeroProb{}, else Beta(\hpFBetaA{}, \hpFBetaB{})\\
factors $m$ & U\{1, \ldots, $\lceil q/\hpFactorFracInv{} \rceil$\}, $q$ the size of the factor block\\
communality: mean, concentration & U(\hpHTwoMeanLo{}, \hpHTwoMeanHi{}), U(\hpHTwoConcLo{}, \hpHTwoConcHi{}), cap \hpHTwoCap{}\\
high band: probability, mean & \hpHTwoHighProb{}, U(\hpHTwoHighMeanLo{}, \hpHTwoHighCap{}), cap \hpHTwoHighCap{}\\
share of $h^2_j$ on other factors & \hpCrossLoadShare{}\\
sign flip of a loading & \hpNegLoadProb{}\\
loading $|w_j|$ of a generic column & U(0, \hpCrossBlockWMax{})\\
marginal types & Gaussian \hpMargGaussian{}, Likert \hpMargLikert{}, binary \hpMargBinary{},\\
 & count \hpMargCount{}, skewed \hpMargSkewed{}\\
binary base rate, Likert levels & U(\hpBinaryRateLo{}, \hpBinaryRateHi{}), \hpLikertLevelsMin{} to \hpLikertLevelsMax{}\\
\midrule
\multicolumn{2}{@{}l}{\emph{Active-effects prior}}\\
heredity & strong\\
probability $\alpha$ of the count prior & \hpCountShare{}\\
count prior, main effects: $\zeta_{\mathrm{main}}$, $\bar{k}_{\mathrm{main}}$ & \hpKMainZero{}, \hpKMainMean{}\\
count prior, interactions: $\zeta_{\mathrm{int}}$, $\bar{k}_{\mathrm{int}}$ & \hpKIntZero{}, \hpKIntMean{}\\
rate prior: $\omega_{\mathrm{main}}$, $\omega_{\mathrm{int}}$ & U(\hpPiMainLo{}, \hpPiMainHi{}), U(\hpPiIntLo{}, \hpPiIntHi{})\\
\midrule
\multicolumn{2}{@{}l}{\emph{Effect-size prior}}\\
slab scale $c$: median $e^{\mu_c}$, $s_c$, truncation & \hpCMedian{}, \hpSC{}, [\hpCMin{}, \hpCMax{}]\\
interaction ratio $\rho$, intercept variance $\tau_0^2$ & \hpRho{}, \hpTauZeroSq{}\\
noise variance $\sigma^2$: $a_0$, $b_0$ & \hpAZero{}, \hpBZero{}\\
\midrule
\multicolumn{2}{@{}l}{\emph{Implied standardized effects, Q1 / median / Q3}}\\
active main effects & \priorMainsPTwoFive{} / \priorMainsPFiveZero{} / \priorMainsPSevenFive{}\\
active interactions & \priorIntPTwoFive{} / \priorIntPFiveZero{} / \priorIntPSevenFive{}\\
meta-analytic norms \citep{gignac2016effect} & \priorNormPTwoFive{} / \priorNormPFiveZero{} / \priorNormPSevenFive{}\\
\bottomrule
\end{tabular*}
\end{table}

\subsection{Predictor prior}
\label{app:predictor-prior}

The predictor matrix is drawn in three steps (Figure~\ref{fig:predictor-prior}). The latent columns are drawn in two blocks, their order is shuffled (Section~\ref{sec:prior}), and a Gaussian copula maps each column to its marginal. Table~\ref{tab:prior-config} lists the fixed values of every step.

Of the $p$ columns, $\mathrm{round}(f\,p)$ form the factor block and the rest form the generic block. The point mass of $f$ at zero keeps predictor matrices without factor structure frequent at every $p$. The columns of the generic block are jointly Gaussian with correlation matrix $R$. The matrix $R$ is a random correlation matrix with a given spectrum, built by Givens rotations \citep{davies2000numerically}, and its eigenvalues are a Dirichlet draw scaled to sum to the block size. Some generic blocks are therefore nearly orthogonal and others nearly singular.

The factor block follows the factor model $\mathbf{F}\Lambda^{\top} + \Psi^{1/2}\mathbf{E}$ with $m$ latent factors. The number $m$ is drawn uniformly between one and a fixed fraction of the block size. Each column $j$ receives a primary factor and a communality $h^2_j$, the share of its variance that the factors explain. The communalities follow a Beta distribution whose mean and concentration are drawn once per dataset, so one dataset has weak and another strong factor structure. A fixed share of $h^2_j$ goes to the primary loading. The remainder is spread over the other factors in Dirichlet-distributed proportions, the cross-loadings. Each loading changes sign with a fixed probability, as reverse-coded items do. The uniqueness is $1 - h^2_j$, so every latent column has unit variance. With a fixed probability the communality mean is drawn from a high band instead. A factor that carries a few columns with communalities near one produces a group of near-duplicate columns. Real predictor matrices contain such groups, and the low band produces them rarely. Each generic column also receives a small loading on a random direction in factor space, because the correlation between the blocks of real data is never exactly zero.

The copula \citep{cario1997modeling} pushes each latent column through the standard normal distribution function and then through the quantile function of a type drawn per column from the Gaussian, Likert, binary, count and skewed families. The parameters of the type (number of levels, base rate, mean count, skew) are drawn per column as well. A column that the transform makes constant draws its type again on the same latent column. Every column is then z-standardized. The interaction columns of $\mathbf{Z}$ are products of the standardized columns and are not standardized again. Discretization attenuates the latent correlations, and skewed marginals produce heavy-tailed interaction columns. A predictor prior with Gaussian columns only would produce neither.

\subsection{The active-effects prior in full}
\label{app:active-set}

\begin{figure}[t]
\begin{center}
\includegraphics[width=\linewidth]{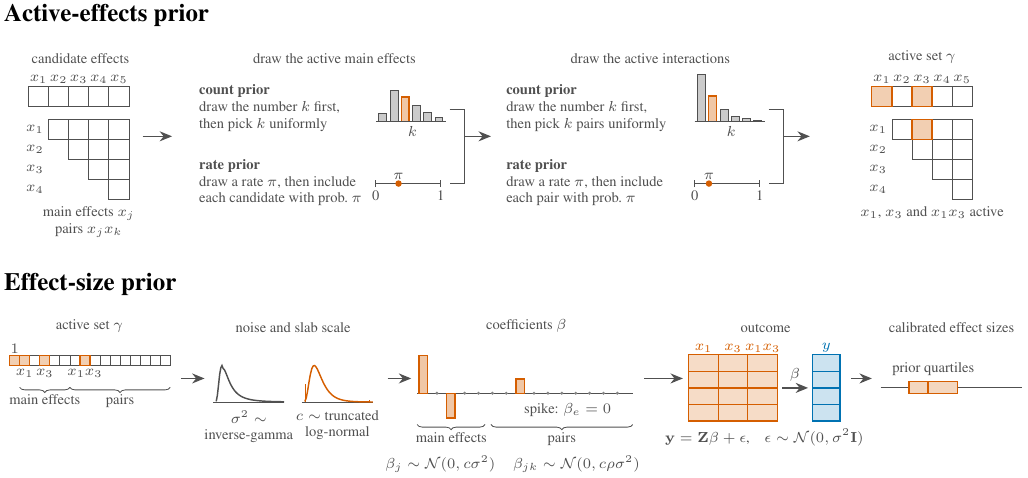}
\end{center}
\caption{The active-effects and effect-size priors, one draw with $p = 5$. Top: the active main effects and then the active interactions are drawn from the candidate set, under the count prior (the number first, then a uniform subset) or the rate prior (a rate first, then one inclusion per candidate). Bottom: the noise variance and the slab scale are drawn, active effects receive slab coefficients and inactive ones a spike at zero, and the outcome is $\mathbf{y} = \mathbf{Z}\beta + \epsilon$.}
\label{fig:active-outcome}
\end{figure}

Figure~\ref{fig:active-outcome} shows one draw of the active set and the outcome. Both components of the active-effects prior first choose the active main effects and then the active interactions among the $M_{\mathrm{int}} = \binom{k_{\mathrm{main}}}{2}$ eligible pairs.

Under the count prior, a count over $M$ candidates is zero with probability $\zeta$. Otherwise it follows a geometric distribution on the positive integers with mean $\bar{k}$, truncated to at most $M$ and renormalized. The main effects use $M = p$, $\zeta_{\mathrm{main}} = \hpKMainZero{}$ and $\bar{k}_{\mathrm{main}} = \hpKMainMean{}$. The interactions use $M = M_{\mathrm{int}}$, $\zeta_{\mathrm{int}} = \hpKIntZero{}$ and $\bar{k}_{\mathrm{int}} = \hpKIntMean{}$, and their count is zero when no pair is eligible. Given the counts, the active main effects and the active interactions are uniform subsets of their candidates, so
\begin{equation*}
\pi_{\mathrm{count}}(\gamma \mid p) = \frac{P_p(k_{\mathrm{main}})}{\binom{p}{k_{\mathrm{main}}}}\,\frac{P_{M_{\mathrm{int}}}(k_{\mathrm{int}})}{\binom{M_{\mathrm{int}}}{k_{\mathrm{int}}}},
\end{equation*}
where $P_M$ is the count distribution over $M$ candidates.

Under the rate prior, an inclusion rate $\omega_{\mathrm{main}}$ is drawn uniformly between \hpPiMainLo{} and \hpPiMainHi{}, and each main effect is active independently with probability $\omega_{\mathrm{main}}$. A second rate $\omega_{\mathrm{int}}$ is drawn uniformly between \hpPiIntLo{} and \hpPiIntHi{}, and each eligible pair is active independently with probability $\omega_{\mathrm{int}}$. Integrating out the two rates gives
\begin{equation*}
\pi_{\mathrm{rate}}(\gamma \mid p) = \E_{\omega_{\mathrm{main}}}\!\big[\omega_{\mathrm{main}}^{k_{\mathrm{main}}}(1-\omega_{\mathrm{main}})^{p-k_{\mathrm{main}}}\big]\,
\E_{\omega_{\mathrm{int}}}\!\big[\omega_{\mathrm{int}}^{k_{\mathrm{int}}}(1-\omega_{\mathrm{int}})^{M_{\mathrm{int}}-k_{\mathrm{int}}}\big],
\end{equation*}
and each expectation over a uniform rate is a difference of incomplete Beta functions.

Each dataset follows the count prior with probability $\alpha = \hpCountShare{}$ and the rate prior otherwise, so $\pi(\gamma \mid p) = \alpha\,\pi_{\mathrm{count}}(\gamma \mid p) + (1-\alpha)\,\pi_{\mathrm{rate}}(\gamma \mid p)$. The exact posterior evaluates this closed form for every model it enumerates. A dataset has no active main effect, and under strong heredity no active effect at all, with probability \priorNullFrac{} under the prior. This share combines the zero mass $\zeta_{\mathrm{main}}$ of the count prior with the probability that the rate prior includes no main effect, which is largest at small $p$ (\priorNullFracPTwo{} at $p = \trainPMin{}$).

\section{Network and training details}
\label{app:model}

\subsection{Architecture}
\label{app:architecture}

Figure~\ref{fig:heads} shows the three heads. The query cells of the outcome column are filled with the mean of the context outcomes before embedding, which hides the query outcomes from the network. Each transformer block has \modelHeads{} attention heads. The bar distribution of the regression head has \modelBins{} bins of equal width on $[\modelBarLo{}, \modelBarHi{}]$ and a half-normal tail on each side \citep{müller2024transformersbayesianinference,hollmann2025accurate}. The main and pair MLPs of the selection and coefficient heads have one hidden layer of \modelMlp{} units each, and they share the pooled vectors of Equation~\ref{eq:pooled}, which are computed once per forward pass. The selection head returns one logit for every main effect and every pair, with no heredity mask. The per-effect heads read the context rows only, and their outputs do not depend on the order of those rows. During pretraining the query rows exist because the predictive term of the loss is scored on held-out rows.

\subsection{Why the predictive distribution needs its own head}
\label{app:predictive}

In a linear model the posterior mean of the coefficients determines the mean of the prediction exactly. For a new row with expanded predictors $\mathbf{z}^{*}$,
\begin{equation*}
\E[y^{*} \mid T] \;=\; (\mathbf{z}^{*})^{\top}\,\E[\beta \mid T],
\end{equation*}
so the coefficient head already implies the predictive mean. The predictive variance does not follow from the posterior means. It decomposes as
\begin{equation*}
\mathrm{Var}(y^{*} \mid T) \;=\; \E[\sigma^2 \mid T] \;+\; (\mathbf{z}^{*})^{\top}\,\mathrm{Cov}(\beta \mid T)\,\mathbf{z}^{*}.
\end{equation*}
The first term is the expected noise variance, the scatter of a new observation around the regression line even if the coefficients were known exactly. It does not shrink as the data grow, while the second term does, along every direction the data inform. The second term is the uncertainty about the regression line at that row. If coefficient vectors were drawn from the posterior, it would be the variance of the predictions $(\mathbf{z}^{*})^{\top}\beta$ they give. Written out,
\begin{equation*}
(\mathbf{z}^{*})^{\top}\,\mathrm{Cov}(\beta \mid T)\,\mathbf{z}^{*}
\;=\; \sum_{e} (z^{*}_e)^2\,\mathrm{Var}(\beta_e \mid T)
\;+\; \sum_{e \neq e'} z^{*}_e z^{*}_{e'}\,\mathrm{Cov}(\beta_e, \beta_{e'} \mid T),
\end{equation*}
so it depends on the covariances between the coefficients as well as on their variances.

\subsection{Training}
\label{app:training}

The datasets of one batch share their size $(n, p)$. The share of query rows is drawn per batch, uniformly between \trainQueryFracLo{} and \trainQueryFracHi{}, and the remaining rows form the context. Muon \citep{jordan2024muon} trains the weight matrices of the backbone and AdamW \citep{loshchilov2019decoupled} every other parameter, at the learning rates of Table~\ref{tab:model-card}. Both follow \trainWarmup{} steps of linear warm-up and then a cosine decay. The gradients of all three loss terms reach the shared backbone. The run was data-parallel on one node. The accumulation steps were divided among the \trainGpuCountWord{} processes, so the effective batch equals that of a single-GPU run. The scheduler requeued the job \trainRestartsWord{}. The run resumed from its last checkpoint, written every \trainCkptMinutes{} minutes, at step \trainResumedStep{}, and the wall time in Table~\ref{tab:model-card} counts every segment.

\begin{table}[t]
\caption{Model card of \textsc{LinearPFN}. Wall time sums the segments of the requeued job, and GPU-hours are wall time times the number of GPUs.}
\label{tab:model-card}
\centering
\footnotesize
\begin{tabular*}{\textwidth}{@{}l@{\hspace{1em}}l@{\extracolsep{\fill}}l@{\extracolsep{0pt}\hspace{1em}}l@{}}
\toprule
layers / heads / embedding & \modelLayers{} / \modelHeads{} / \modelEmb{} & parameters & \modelParams{} (\modelParamsExact{})\\
MLP hidden units & \modelMlp{} & GPUs & \trainGpuCount{} $\times$ \trainGpu{}\\
optimizer steps & \trainSteps{} & wall time & \trainWallHours{} h\\
effective batch & \trainEffBatch{} (\trainPhysBatch{} $\times$ \trainAccum{}) & GPU-hours & \trainGpuHours{}\\
datasets seen & \trainDatasetsSeen{} & throughput & \trainStepsPerSec{} steps/s\\
learning rate & Muon \trainMuonLr{}, AdamW \trainLr{} & peak GPU memory & \trainPeakMemGiB{} GiB\\
\bottomrule
\end{tabular*}
\end{table}


\section{The exact posterior and the MCMC sampler}
\label{app:exact}

\subsection{Enumeration and quadrature}
\label{app:enumeration}

Given $(\gamma, c)$, the prior of Equation~\ref{eq:prior} is a normal-inverse-gamma regression on the active columns $\mathbf{Z}_\gamma$. The coefficients follow $\beta_\gamma \mid \sigma^2 \sim \gN(0, \sigma^2 V_0)$, where $V_0$ is the diagonal matrix of the slab variances $v_e$ of the active effects, and $\sigma^2 \sim \mathrm{InvGamma}(a_0, b_0)$. The marginal likelihood is
\begin{equation}
\log p(\mathbf{y}\mid\gamma,c) = \frac12\log\frac{|V_n|}{|V_0|} + \log\frac{\Gamma(a_n)}{\Gamma(a_0)} + a_0\log b_0 - a_n\log b_n - \frac{n}{2}\log 2\pi,
\label{eq:evidence}
\end{equation}
with the posterior quantities
\begin{align*}
V_n &= \bigl(V_0^{-1} + \mathbf{Z}_\gamma^\top\mathbf{Z}_\gamma\bigr)^{-1}, & m_n &= V_n \mathbf{Z}_\gamma^\top\mathbf{y}, \\
a_n &= a_0 + \tfrac{n}{2}, & b_n &= b_0 + \tfrac12\bigl(\mathbf{y}^\top\mathbf{y} - m_n^\top\mathbf{Z}_\gamma^\top\mathbf{y}\bigr).
\end{align*}
Because $\rho$ is fixed, $c$ is the only continuous hyperparameter, and the evidence of a model is
\begin{equation*}
p(\mathbf{y}\mid\gamma) = \int p(\mathbf{y}\mid\gamma,c)\,p(c)\,dc.
\end{equation*}
The integral is computed by quadrature.

The reference enumerates every model with nonzero prior mass. These are the models with the intercept active in which every active interaction has both parents active. At $p = \exactMaxP{}$, the largest $p$ the reference enumerates, there are \exactNModelsMaxP{} of them. The posterior model probabilities are
\begin{equation*}
w_\gamma \propto \pi(\gamma \mid p)\,p(\mathbf{y}\mid\gamma),
\end{equation*}
with the prior of Appendix~\ref{app:active-set}. They give the exact PIPs as $\sum_\gamma w_\gamma \gamma$ and the posterior-mean coefficients as the average of $m_n$ over models and quadrature nodes, with zeros outside each model. For a new row with expanded predictors $\mathbf{z}^*$, each pair of a model and a node contributes a Student-$t$ distribution with $2a_n$ degrees of freedom,
\begin{equation*}
t_{2a_n}\Bigl((\mathbf{z}^*_\gamma)^\top m_n,\ \frac{b_n}{a_n}\bigl(1 + (\mathbf{z}^*_\gamma)^\top V_n\,\mathbf{z}^*_\gamma\bigr)\Bigr),
\end{equation*}
with its location and squared scale as arguments. The exact posterior predictive distribution is the mixture of these components, and its quantiles are found by bisection on the mixture distribution function.

The integral over $c$ uses \quadNodesC{} Gauss--Legendre nodes on $\log c$ over $[\log c_{\min}, \log c_{\max}]$. The mass of a node is its quadrature weight times the Jacobian times the truncated lognormal density of $c$, and the masses are renormalized to sum to one. The posterior is then a discrete distribution over pairs of a model and a node. The computation runs in two passes. The first pass computes the evidence of every model at every node and sums over the nodes, so the model probabilities and the PIPs use every model. The second pass computes $m_n$, $V_n$ and $b_n$ only for the pairs that together carry all but \quadPruneTol{} of the posterior mass, and the coefficient means and the predictive distribution use these pairs.

The node count was checked against a grid of \quadFinestCInt{} nodes on \quadNDatasets{} prior draws in each cell that crosses \quadCellsNInt{} rows with \quadCellsPInt{} predictors. At \quadNodesC{} nodes the largest deviation occurs at $n = \quadChosenWorstNInt{}$ and $p = \quadChosenWorstPInt{}$. It is \quadChosenMaxDPipInt{} in the PIPs and \quadChosenMaxDLogEvInt{} in the log evidence of a model. Over the tested grids of \quadCGridInt{} nodes, the largest PIP deviation falls from \quadByCMaxDPipIntOneTwo{} to \quadChosenMaxDPipInt{}.

\subsection{The MCMC sampler}
\label{app:mcmc}

Beyond the enumeration reach the reference is an MCMC sampler over the model space \citep{madigan1995bayesian} that targets the same posterior. It is a Metropolis--Hastings chain on the active set with moves that add, drop or swap one effect. Its target is $\pi(\gamma \mid p)\,p(\mathbf{y}\mid\gamma)$ with the evidence of Equation~\ref{eq:evidence} and the quadrature of Appendix~\ref{app:enumeration}, so models that violate strong heredity have zero target mass. Several chains start from prior draws, and the burn-in states are discarded. PIPs are visit frequencies, and the posterior means average the posterior mean of each visited model. A run meets the convergence criteria when the largest split $\hat R$ is below \realyRhatThreshold{} and the smallest effective sample size is above \realyEssThreshold{}.


\section{Validation against the exact posterior}
\label{app:validation}

Networks trained on simulated data to return a posterior over model structure also exist for scientific simulators \citep{schroeder2024sbmi,gloeckler2026scalable}, but a simulator whose likelihood cannot be evaluated offers no exact posterior to compare with. The conjugate prior of \textsc{LinearPFN} provides one for a small number of predictors.

The validation panel consists of prior draws. It crosses \valNValues{} context rows with \valPValues{} predictors, with \valDatasetsPerCell{} datasets per cell and \valNQuery{} query rows per dataset. In the cells with $p \le \exactMaxP{}$, the exact cells, the reference of Appendix~\ref{app:exact} is computed on the same datasets. The coverage criterion pools every cell of the panel, and the other five criteria use the exact cells.

\begin{enumerate}
\item \emph{Predictive headroom.} In each exact cell, $(\text{NLL}_{\mathrm{marginal}} - \text{NLL}_{\mathrm{model}}) / (\text{NLL}_{\mathrm{marginal}} - \text{NLL}_{\mathrm{exact}})$ is at least \critHeadroomMin{}. Here the marginal is a Gaussian fitted to the context outcomes, and each NLL is the mean over the query rows. Cells in which the exact posterior improves on the marginal by less than \critHeadroomFlagNats{} nats are reported and not scored, because the ratio of two small differences is unstable. The worst scored cell counts.
\item \emph{Coefficient recovery.} The correlation between the coefficient head's posterior means and the exact posterior means over the candidate effects, pooled over the exact cells, is at least \critCoefRMin{}. The probe is reported beside it and not scored. It reaches \finalCoefRProbeExactPanel{} on the same datasets.
\item \emph{Coverage.} The central predictive intervals at the nominal levels \critCoverageLevels{} cover the query outcomes within \critCoverageTolPp{} percentage points of nominal.
\item \emph{Selection ranking.} The AUC of the PIPs against the simulated active set is at most \critAucGap{} below the AUC of the exact posterior on the same datasets, in the worst of the pooled, main-effect and interaction subsets.
\item \emph{Selection calibration.} The ECE of the PIPs exceeds the ECE of the exact posterior by at most \critEceGap{}, in the worst subset.
\item \emph{Selection resolution.} The resolution term of the Murphy decomposition of the PIPs is at least \critResPooled{} times that of the exact posterior on the pooled effects and at least \critResSubset{} times on each subset.
\end{enumerate}
Resolution alone can be raised by extreme PIPs at the cost of calibration, so it is scored together with the ECE.

The probe reads coefficients from the predictive distribution. It evaluates the mean of the predictive distribution at \probePointsPerEffect{}$\,d$ probe points, which are context rows resampled with replacement plus Gaussian jitter with standard deviation \probeJitterSd{} on each standardized predictor. It then regresses these means on the columns of $\mathbf{Z}$ by ridge regression with penalty \probeRidge{}. Under the exact posterior the predictive mean is linear in $\mathbf{Z}$ (Appendix~\ref{app:predictive}), so the probe recovers the exact posterior means. 


\section{Benchmark setup}
\label{app:benchmark}

\subsection{The real predictor matrices}
\label{app:predictor-matrices}

The predictor matrices come from the \textsc{Rdatasets} collection \citep{Rdatasets,Rlang}, an index of \rxIndexN{} datasets shipped with \rxIndexPackages{} R packages. A hand-made label file assigns each package a domain, and \rxIncludedPackages{} packages are labeled as social-science or psychology data. Their datasets form the frame of \rxFrameN{} tables, after tables described as simulated, tables stored as time series or contingency tables, and tables with fewer than \rxMinN{} rows or \rxMinP{} usable columns in the index are set aside.

\begin{table}[t]
\caption{\textsc{Rdatasets} entries that did not become a predictor matrix, by reason.}
\label{tab:real-x-exclusions}
\centering
\footnotesize
\begin{tabular*}{\textwidth}{@{}l@{\extracolsep{\fill}}rlr@{}}
\toprule
reason & entries & reason & entries\\
\midrule
package outside the frame & \rxExclPackageExcluded{} & described as simulated & \rxExclDescriptionKeyword{}\\
index: fewer than \rxMinN{} rows & \rxExclIndexRowsLtMin{} & fewer than \rxMinP{} columns after cleaning & \rxExclRealizedPLtMin{}\\
index: fewer than \rxMinP{} usable columns & \rxExclIndexPModelLtMin{} & more than \rxMaxRowLoss{} of rows lost & \rxExclRowLossGtMax{}\\
time-series or contingency-table object & \rxExclObjectClass{} & same title as another table & \rxExclDuplicateTitle{}\\
pure time series & \rxExclPureTimeSeries{} & download failed & \rxExclDownloadFailed{}\\
same columns as a larger table & \rxExclDuplicateColumns{} & fewer than \rxMinN{} rows after cleaning & \rxExclRealizedNLtMin{}\\
\midrule
\multicolumn{3}{@{}l}{total} & \rxExclTotal{}\\
\bottomrule
\end{tabular*}
\end{table}

One cleaning function with no per-dataset switches processes every table of the frame. It drops identifier columns. It admits numeric, logical and two-level columns, the last two coded 0/1, and drops categorical columns with more than two levels. It drops columns with more than \rxMaxColMissing{} missing values and keeps the complete cases. A table that loses more than \rxMaxRowLoss{} of its rows in this step is excluded. The function then drops constant columns, every column whose absolute correlation with an earlier kept column is at least \rxMaxAbsCor{}, and every column with fewer than \rxMinMinorityRows{} rows outside its most frequent value. A table is excluded if fewer than \rxMinN{} rows or \rxMinP{} columns remain. Pure time series and duplicate tables are removed. This leaves \rxEligibleN{} eligible tables, and each of them is one predictor matrix of the main benchmark and of its variants with outcomes outside the prior. A table with more than \rxPoolNMax{} rows or \rxPoolPMax{} predictors is subsampled once with a seed derived from its name, in rows for \rxPoolRowSubsampled{} tables and in columns for \rxPoolColSubsampled{}. The rule on near-constant columns is applied again after the subsampling. Every table that does not become a predictor matrix carries a logged reason (Table~\ref{tab:real-x-exclusions}).

The \rxPoolN{} predictor matrices come from \rxPoolPackages{} packages in \rxPoolDomainsN{} fields: economics (\rxPoolDomainEconomics{} predictor matrices), political science (\rxPoolDomainPoliticalScience{}), general social science (\rxPoolDomainSocialScience{}), education (\rxPoolDomainEducation{}), criminology (\rxPoolDomainCriminology{}), psychology (\rxPoolDomainPsychology{}) and demography (\rxPoolDomainDemography{}). Their median numbers of rows and predictors are \rxPoolNMedian{} and \rxPoolPMedian{}, and \rxPoolPLeFive{} of them have at most \exactMaxP{} predictors. The correlation rule acts before the subsampling, so a subsampled predictor matrix can exceed it. The largest absolute correlation between two columns of a predictor matrix has median \rxPoolMaxAbsRMedian{} and maximum \rxPoolMaxAbsRMax{}. In \rxPoolRowUnitPanel{} of the predictor matrices the rows are repeated observations of the same units.

\subsection{Construction of the main benchmark}
\label{app:panel-construction}

\begin{figure}[t]
\begin{center}
\includegraphics[width=\linewidth]{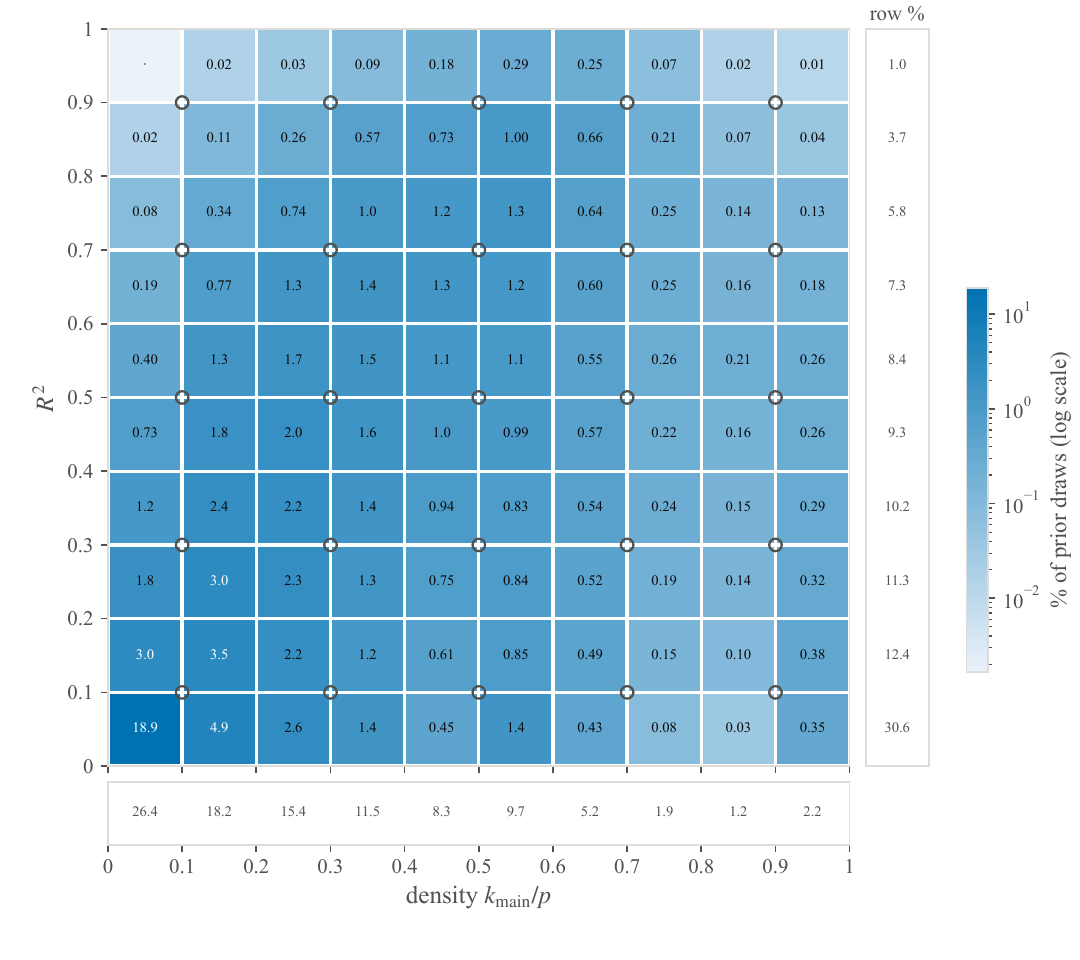}
\end{center}
\caption{Share of \priorGridDatasets{} datasets drawn from the prior in each cell of the plane of density and $R^2$, on a log color scale, with the row and column totals in the margins. Rings mark the lower-left corners of the \ipCellsRealX{} cells of the main benchmark, which together hold \priorGridBenchCellsTotalPct{}\,\% of the draws. A dot marks a cell whose share rounds to zero.}
\label{fig:prior-grid}
\end{figure}

Every dataset of the main benchmark is accepted by rejection sampling. A draw picks a predictor matrix uniformly from the \rxPoolN{} predictor matrices and draws $(\gamma, c, \sigma^2, \beta, \mathbf{y})$ from the prior of Appendix~\ref{app:prior} on its standardized columns. Density and $R^2$ are defined as in Section~\ref{sec:eval}. Each axis of the grid has \ipLevelsRealX{} bins of width \ipBinWidthRealX{}, with equally spaced lower edges from \ipLevelLoRealX{} to \ipLevelHiRealX{}, and the top bin includes one. The bins do not touch. A draw whose density or $R^2$ lies between two bins or below the lowest bin is discarded, so the benchmark contains no dataset without an active main effect and none with an $R^2$ below the lowest bin. The \ipCellsRealX{} cells hold \priorGridBenchCellsTotalPct{}\,\% of the prior mass (Figure~\ref{fig:prior-grid}), and within them every cell is weighted equally. Each cell has its own random stream and draws until it is full. The share of accepted draws ranges from \ipAcceptMaxPctRealX{} in the most frequent cell to \ipAcceptMinPctRealX{} in the rarest, because the prior rarely produces dense or high-$R^2$ datasets (Figure~\ref{fig:prior-grid}).

The density $k_{\mathrm{main}}/p_\ell$ takes only multiples of $1/p_\ell$, so a predictor matrix with few predictors reaches only some density bins. None of the \ipDatasetsExactRealX{} datasets with $p_\ell \le \exactMaxP{}$ lies in the lowest density bin, because $1/p_\ell$ is at least the upper edge of that bin. Every predictor matrix appears in at least one cell, and a cell holds between \ipDesignsPerCellMinRealX{} and \ipDesignsPerCellMaxRealX{} distinct predictor matrices. The cells therefore differ in the predictor matrices they hold, and a comparison across cells mixes the effect of density and $R^2$ with the effect of those predictor matrices.

\subsection{Methods}
\label{app:method-set}

The lasso \citep{tibshirani1996regression} selects the effects that are nonzero at the penalty with the smallest cross-validated error, and it ranks effects by their point of entry on the regularization path. Stability selection with complementary pairs \citep{meinshausenStabilitySelection2010,shah2013variable} fits the lasso path on \stabPairs{} complementary pairs of half-samples. It selects the effects whose selection frequency is at least \stabThreshold{} and ranks by that frequency. It returns no coefficients of its own, so its coefficients are the least-squares fit on its selected effects, and its fit seconds include that fit. \textsc{SuSiE} \citep{susieR1} with at most \susieL{} single effects selects the effects with a PIP above one half and ranks by PIP. \textsc{glinternet} \citep{gli2026} and \textsc{hierNet} \citep{hier2020} select the effects at the penalty with the smallest cross-validated error under their heredity constraints, and they rank by the point of entry on the path. The hard-zeroed read-out of \textsc{LinearPFN} (Section~\ref{sec:results}) keeps the intercept and involves no refit.

The coefficients of the lasso, \textsc{glinternet} and \textsc{hierNet} are penalized estimates. The exact posterior, like \textsc{LinearPFN}, returns a posterior mean over models, in which no coefficient is exactly zero.

\subsection{Metrics}
\label{app:metrics}

The metrics are those of Section~\ref{sec:eval}, with the following details. The AUC of a dataset ranks the candidate effects by the method's own score, the PIP for \textsc{LinearPFN} and \textsc{SuSiE}, the selection frequency for stability selection and the point of entry on the path for the lasso, \textsc{glinternet} and \textsc{hierNet}. Tied scores receive their mean rank. The AUC is undefined on a dataset in which every candidate effect is active or none is, so the AUC average covers \ipAucNOursRealX{} of the \ipDatasetsOursRealX{} datasets. The coefficient RMSE excludes the intercept. A paired difference is \textsc{LinearPFN} minus the other method on the datasets that both score. Its interval is a seeded bootstrap 95\,\% interval of the mean difference over \bootReps{} resamples, and its $p$ value is that of the Wilcoxon signed-rank test. Fit time is the wall time of one fit on one CPU thread of the cluster, measured inside each method's own process around the fit and its tuning.


\section{Benchmark results}
\label{app:bench-results}

\subsection{Main benchmark: full results}
\label{app:main-bench}

Table~\ref{tab:ip-paired} gives the paired differences between \textsc{LinearPFN} and each baseline. The means and quartiles of every metric are in Table~\ref{tab:ip-summary}.

\begin{table}[t]
\caption{Main benchmark: paired differences, \textsc{LinearPFN} minus each baseline on the same datasets, with bootstrap 95\,\% intervals. The last column pairs the hard-zeroed read-out of \textsc{LinearPFN} with each baseline. Every Wilcoxon signed-rank $p$ is below $0.001$, except for the RMSE against \textsc{hierNet} ($p \ipPRmseHierNetRealX{}$) and the hard-zeroed RMSE against \textsc{glinternet} ($p \ipHzPRmseGlinternetRealX{}$).}
\label{tab:ip-paired}
\centering
\footnotesize
\setlength{\tabcolsep}{3pt}
\begin{tabular*}{\textwidth}{@{}l@{\extracolsep{\fill}}rrrr@{}}
\toprule
baseline & $F_1$/ds & AUC/ds & coefficient RMSE & RMSE, hard-zeroed\\
\midrule
\textsc{hierNet} & \ipDiffFOneHierNetRealX{} & \ipDiffAucHierNetRealX{} & \ipDiffRmseHierNetRealX{} & \ipHzDiffRmseHierNetRealX{}\\
 & \iqr{\ipDiffLoFOneHierNetRealX{}}{\ipDiffHiFOneHierNetRealX{}} & \iqr{\ipDiffLoAucHierNetRealX{}}{\ipDiffHiAucHierNetRealX{}} & \iqr{\ipDiffLoRmseHierNetRealX{}}{\ipDiffHiRmseHierNetRealX{}} & \iqr{\ipHzDiffLoRmseHierNetRealX{}}{\ipHzDiffHiRmseHierNetRealX{}}\\[2pt]
\textsc{glinternet} & \ipDiffFOneGlinternetRealX{} & \ipDiffAucGlinternetRealX{} & \ipDiffRmseGlinternetRealX{} & \ipHzDiffRmseGlinternetRealX{}\\
 & \iqr{\ipDiffLoFOneGlinternetRealX{}}{\ipDiffHiFOneGlinternetRealX{}} & \iqr{\ipDiffLoAucGlinternetRealX{}}{\ipDiffHiAucGlinternetRealX{}} & \iqr{\ipDiffLoRmseGlinternetRealX{}}{\ipDiffHiRmseGlinternetRealX{}} & \iqr{\ipHzDiffLoRmseGlinternetRealX{}}{\ipHzDiffHiRmseGlinternetRealX{}}\\[2pt]
SuSiE & \ipDiffFOneSusieRealX{} & \ipDiffAucSusieRealX{} & \ipDiffRmseSusieRealX{} & \ipHzDiffRmseSusieRealX{}\\
 & \iqr{\ipDiffLoFOneSusieRealX{}}{\ipDiffHiFOneSusieRealX{}} & \iqr{\ipDiffLoAucSusieRealX{}}{\ipDiffHiAucSusieRealX{}} & \iqr{\ipDiffLoRmseSusieRealX{}}{\ipDiffHiRmseSusieRealX{}} & \iqr{\ipHzDiffLoRmseSusieRealX{}}{\ipHzDiffHiRmseSusieRealX{}}\\[2pt]
lasso & \ipDiffFOneLassoRealX{} & \ipDiffAucLassoRealX{} & \ipDiffRmseLassoRealX{} & \ipHzDiffRmseLassoRealX{}\\
 & \iqr{\ipDiffLoFOneLassoRealX{}}{\ipDiffHiFOneLassoRealX{}} & \iqr{\ipDiffLoAucLassoRealX{}}{\ipDiffHiAucLassoRealX{}} & \iqr{\ipDiffLoRmseLassoRealX{}}{\ipDiffHiRmseLassoRealX{}} & \iqr{\ipHzDiffLoRmseLassoRealX{}}{\ipHzDiffHiRmseLassoRealX{}}\\[2pt]
stability selection & \ipDiffFOneStabilityRefitRealX{} & \ipDiffAucStabilityRefitRealX{} & \ipDiffRmseStabilityRefitRealX{} & \ipHzDiffRmseStabilityRefitRealX{}\\
 & \iqr{\ipDiffLoFOneStabilityRefitRealX{}}{\ipDiffHiFOneStabilityRefitRealX{}} & \iqr{\ipDiffLoAucStabilityRefitRealX{}}{\ipDiffHiAucStabilityRefitRealX{}} & \iqr{\ipDiffLoRmseStabilityRefitRealX{}}{\ipDiffHiRmseStabilityRefitRealX{}} & \iqr{\ipHzDiffLoRmseStabilityRefitRealX{}}{\ipHzDiffHiRmseStabilityRefitRealX{}}\\[2pt]
\bottomrule
\end{tabular*}
\end{table}

The \rxPoolRowUnitPanel{} panel predictor matrices contribute \ipUnitNPanelRealX{} datasets. The simulated outcomes are independent across rows on every predictor matrix, so the panel structure enters only through the predictors. \textsc{LinearPFN}'s $F_1$ is \ipUnitFOneMeanOursPanelRealX{} on the panels and \ipUnitFOneMeanOursCrossSectionRealX{} on the cross-sections. Its margin over \textsc{hierNet} is \ipUnitDiffFOneHierNetPanelRealX{} and \ipUnitDiffFOneHierNetCrossSectionRealX{}, and over \textsc{SuSiE} \ipUnitDiffFOneSusiePanelRealX{} and \ipUnitDiffFOneSusieCrossSectionRealX{}.

On one \fitGpuName{} GPU the forward pass takes \fitSecGpu{} seconds at the median over the same \fitSecGpuN{} datasets, \fitSecGpuPNinety{} at the 90th percentile and \fitSecGpuMax{} at most. This is \fitCpuOverGpu{} times faster than the single-thread median, with a one-off \fitSecGpuLoad{} seconds to load the checkpoint. The exact posterior takes \ipFitMedExactRealX{} seconds at the median on the \ipDatasetsExactRealX{} datasets it reaches. The MCMC sampler takes \ipFitMedMcThreeRealX{} seconds at the median and \ipFitPNinetyMcThreeRealX{} at the 90th percentile on its \ipMcmcNRealX{} datasets, \ipFitRatioMcThreeRealX{} times the single-thread median of \textsc{LinearPFN} on the same datasets.

\subsection{Distance to the exact posterior and to MCMC}
\label{app:exact-fidelity}

On the \ipDatasetsExactRealX{} datasets of the main benchmark with $p_\ell \le \exactMaxP{}$, the exact posterior is computed on the same data. \textsc{LinearPFN} minus the exact posterior is \ipDiffAucExactRealX{} [\ipDiffLoAucExactRealX{}, \ipDiffHiAucExactRealX{}] in AUC per dataset and \ipDiffFOneExactRealX{} [\ipDiffLoFOneExactRealX{}, \ipDiffHiFOneExactRealX{}] in $F_1$ per dataset. Its coefficient RMSE exceeds that of the exact posterior by \ipDiffRmseExactRealX{} [\ipDiffLoRmseExactRealX{}, \ipDiffHiRmseExactRealX{}]. This excess is the amortization error in the coefficients. The RMSE of the exact posterior itself, \ipRmseMeanExactRealX{}, is the error that remains with exact inference under the prior.

Beyond the enumeration reach the reference is the MCMC sampler of Appendix~\ref{app:mcmc}. It ran on \ipMcmcNRealX{} datasets of the main benchmark drawn at random over all $p_\ell$, for \ipMcmcStepsMedRealX{} steps per chain at the median. \ipMcmcConvPctRealX{} of the runs meet its convergence criteria, with a median largest split $\hat R$ of \ipMcmcRhatMedRealX{} and a median smallest effective sample size of \ipMcmcEssMedRealX{}. \textsc{LinearPFN} minus the MCMC sampler is \ipDiffAucMcThreeRealX{} [\ipDiffLoAucMcThreeRealX{}, \ipDiffHiAucMcThreeRealX{}] in AUC, \ipDiffFOneMcThreeRealX{} [\ipDiffLoFOneMcThreeRealX{}, \ipDiffHiFOneMcThreeRealX{}] in $F_1$ and \ipDiffRmseMcThreeRealX{} [\ipDiffLoRmseMcThreeRealX{}, \ipDiffHiRmseMcThreeRealX{}] in RMSE, over all \ipPairedNFOneMcThreeRealX{} datasets.

\subsection{Outcomes outside the prior}
\label{app:knobs}

\begin{figure}[p]
\centering
\includegraphics[width=\textwidth]{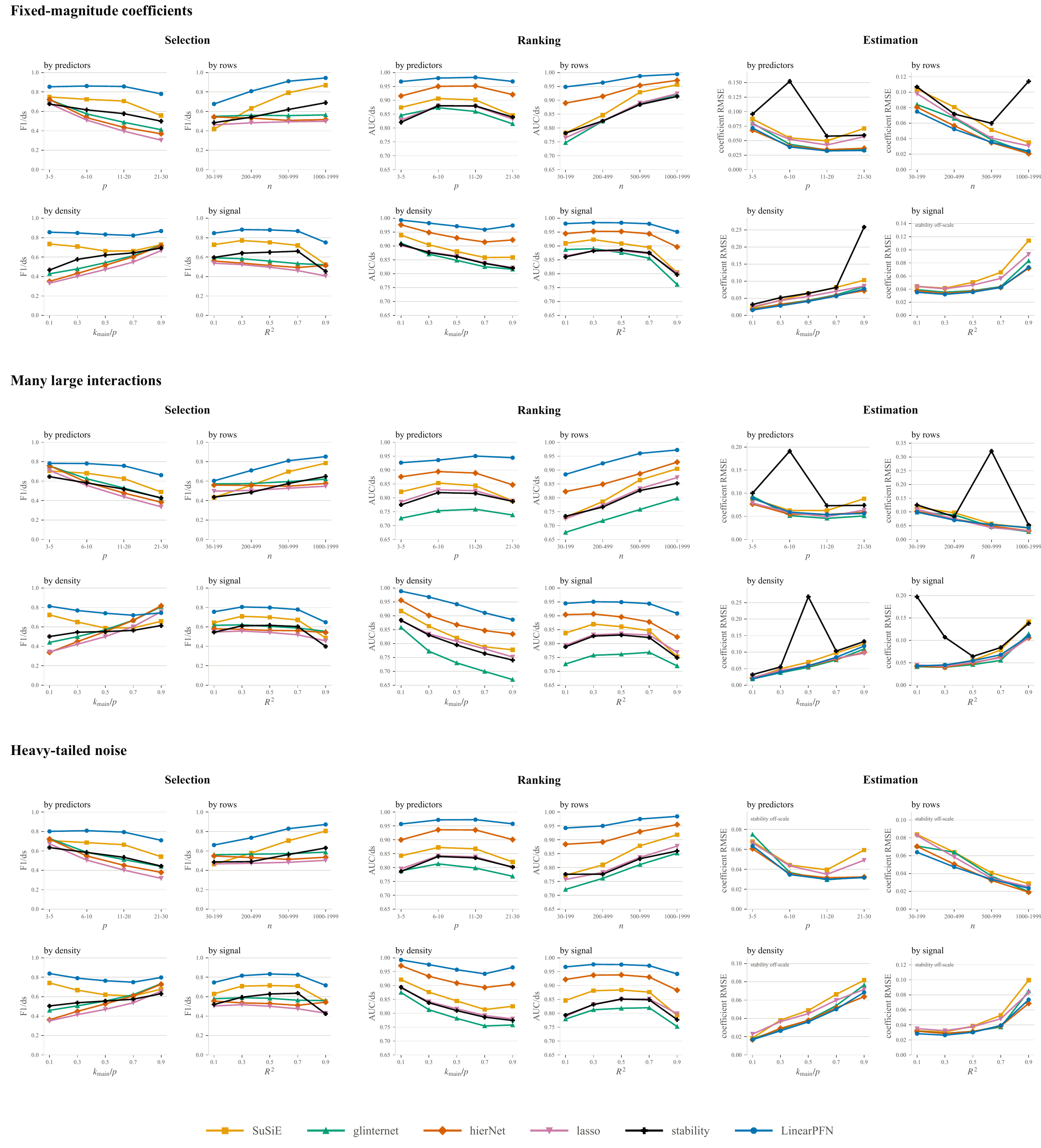}
\caption{Outcomes outside the prior, one row per variant, each on \knobDatasetsOursCoef{} datasets. Left: selection, $F_1$ per dataset at each method's own selection rule. Middle: ranking, AUC per dataset. Right: estimation, coefficient RMSE over every candidate effect. Each panel slices the datasets by the number of predictors, the number of rows, the density $k_{\mathrm{main}}/p$ and the $R^2$; every point is the mean over the datasets of that slice. Density and $R^2$ are the bins of the main-benchmark dataset that the variant is built on, so the $R^2$ of a variant dataset can lie outside its bin. A method with an off-scale value in a panel is left out of that panel, and the panel names it.}
\label{fig:knob-sweeps}
\end{figure}

The variants are built on the datasets of the main benchmark (Appendix~\ref{app:panel-construction}). For each dataset the draw of the main benchmark is replayed from its random stream and checked against the stored dataset, which recovers $c$ and $\sigma^2$, and each variant then applies its change from a random stream of its own. Under fixed-magnitude coefficients every active coefficient keeps the second moment of its slab, and no active effect lies near zero. Under many large interactions the median dataset has \knobIntCountMedianInt{} active interactions (mean \knobIntCountMeanInt{}, maximum \knobIntCountMaxInt{}), and \knobIntNoneShareInt{} of the datasets have none. The density $k_{\mathrm{main}}/p$ of every dataset is the same under all three variants. \textsc{LinearPFN} and the exact posterior assume the prior of Section~\ref{sec:prior}, so on the three variants the exact posterior is the posterior under a wrong prior, and its distance from the truth is the cost of the prior itself. The ECE is computed over every candidate effect of the panel and is reported for the methods that return PIPs. Every method returned a fit on every dataset, so every panel holds \knobDatasetsOursCoef{} datasets. Table~\ref{tab:knobs} lists every method on every panel with its paired difference from the control. Figure~\ref{fig:knob-sweeps} slices each variant by the number of predictors, the number of rows, the density and the $R^2$, as Figure~\ref{fig:ip-sweeps} slices the main benchmark.

The grid truncates and reweights the prior, and its cells hold \priorGridBenchCellsTotalPct{}\,\% of the prior mass (Figure~\ref{fig:prior-grid}). Even the exact posterior is therefore not calibrated on the benchmark, and the ECE here compares methods without testing calibration. The ECE of \textsc{LinearPFN} is \knobEceOursNone{} on the control, against \knobEceSusieNone{} for \textsc{SuSiE}, and \knobEceOursCoef{}, \knobEceOursInt{} and \knobEceOursNoise{} under fixed-magnitude coefficients, many large interactions and heavy-tailed noise. On the datasets within the enumeration reach, the ECE of the exact posterior is \knobExactEceExactNone{} on the control, \knobExactEceExactCoef{} under fixed-magnitude coefficients and \knobExactEceExactInt{} under many large interactions, against \knobExactEceOursNone{}, \knobExactEceOursCoef{} and \knobExactEceOursInt{} for \textsc{LinearPFN} on the same datasets. The $F_1$ of \textsc{LinearPFN} on these datasets is \knobExactFOneOursNone{} against \knobExactFOneExactNone{} for the exact posterior on the control and \knobExactFOneOursInt{} against \knobExactFOneExactInt{} under many large interactions.

\begin{table}[t]
\caption{Outcomes outside the prior: every method on the control (the main benchmark) and on the three variants, means over datasets, with the paired change from the control in parentheses. AUC and $F_1$ per dataset, coefficient RMSE over every candidate effect.}
\label{tab:knobs}
\centering
\footnotesize
\setlength{\tabcolsep}{3pt}
\begin{tabular*}{\textwidth}{@{}l@{\extracolsep{\fill}}rrrrrr@{}}
\toprule
& \multicolumn{3}{c}{control} & \multicolumn{3}{c}{fixed-magnitude coefficients}\\
\cmidrule(lr){2-4}\cmidrule(l){5-7}
method & AUC/ds & $F_1$/ds & RMSE & AUC/ds & $F_1$/ds & RMSE\\
\midrule
\textsc{LinearPFN} & \knobAucOursNone{} & \knobFOneOursNone{} & \knobRmseOursNone{} & \knobAucOursCoef{} {\scriptsize(\knobCtrlDiffAucOursCoef{})} & \knobFOneOursCoef{} {\scriptsize(\knobCtrlDiffFOneOursCoef{})} & \knobRmseOursCoef{} {\scriptsize(\knobCtrlDiffRmseOursCoef{})}\\
\textsc{hierNet} & \knobAucHierNetNone{} & \knobFOneHierNetNone{} & \knobRmseHierNetNone{} & \knobAucHierNetCoef{} {\scriptsize(\knobCtrlDiffAucHierNetCoef{})} & \knobFOneHierNetCoef{} {\scriptsize(\knobCtrlDiffFOneHierNetCoef{})} & \knobRmseHierNetCoef{} {\scriptsize(\knobCtrlDiffRmseHierNetCoef{})}\\
\textsc{glinternet} & \knobAucGlinternetNone{} & \knobFOneGlinternetNone{} & \knobRmseGlinternetNone{} & \knobAucGlinternetCoef{} {\scriptsize(\knobCtrlDiffAucGlinternetCoef{})} & \knobFOneGlinternetCoef{} {\scriptsize(\knobCtrlDiffFOneGlinternetCoef{})} & \knobRmseGlinternetCoef{} {\scriptsize(\knobCtrlDiffRmseGlinternetCoef{})}\\
SuSiE & \knobAucSusieNone{} & \knobFOneSusieNone{} & \knobRmseSusieNone{} & \knobAucSusieCoef{} {\scriptsize(\knobCtrlDiffAucSusieCoef{})} & \knobFOneSusieCoef{} {\scriptsize(\knobCtrlDiffFOneSusieCoef{})} & \knobRmseSusieCoef{} {\scriptsize(\knobCtrlDiffRmseSusieCoef{})}\\
lasso & \knobAucLassoNone{} & \knobFOneLassoNone{} & \knobRmseLassoNone{} & \knobAucLassoCoef{} {\scriptsize(\knobCtrlDiffAucLassoCoef{})} & \knobFOneLassoCoef{} {\scriptsize(\knobCtrlDiffFOneLassoCoef{})} & \knobRmseLassoCoef{} {\scriptsize(\knobCtrlDiffRmseLassoCoef{})}\\
stability selection & \knobAucStabilityRefitNone{} & \knobFOneStabilityRefitNone{} & \knobRmseStabilityRefitNone{} & \knobAucStabilityRefitCoef{} {\scriptsize(\knobCtrlDiffAucStabilityRefitCoef{})} & \knobFOneStabilityRefitCoef{} {\scriptsize(\knobCtrlDiffFOneStabilityRefitCoef{})} & \knobRmseStabilityRefitCoef{} {\scriptsize(\knobCtrlDiffRmseStabilityRefitCoef{})}\\
\midrule
& \multicolumn{3}{c}{many large interactions} & \multicolumn{3}{c}{heavy-tailed noise}\\
\cmidrule(lr){2-4}\cmidrule(l){5-7}
method & AUC/ds & $F_1$/ds & RMSE & AUC/ds & $F_1$/ds & RMSE\\
\midrule
\textsc{LinearPFN} & \knobAucOursInt{} {\scriptsize(\knobCtrlDiffAucOursInt{})} & \knobFOneOursInt{} {\scriptsize(\knobCtrlDiffFOneOursInt{})} & \knobRmseOursInt{} {\scriptsize(\knobCtrlDiffRmseOursInt{})} & \knobAucOursNoise{} {\scriptsize(\knobCtrlDiffAucOursNoise{})} & \knobFOneOursNoise{} {\scriptsize(\knobCtrlDiffFOneOursNoise{})} & \knobRmseOursNoise{} {\scriptsize(\knobCtrlDiffRmseOursNoise{})}\\
\textsc{hierNet} & \knobAucHierNetInt{} {\scriptsize(\knobCtrlDiffAucHierNetInt{})} & \knobFOneHierNetInt{} {\scriptsize(\knobCtrlDiffFOneHierNetInt{})} & \knobRmseHierNetInt{} {\scriptsize(\knobCtrlDiffRmseHierNetInt{})} & \knobAucHierNetNoise{} {\scriptsize(\knobCtrlDiffAucHierNetNoise{})} & \knobFOneHierNetNoise{} {\scriptsize(\knobCtrlDiffFOneHierNetNoise{})} & \knobRmseHierNetNoise{} {\scriptsize(\knobCtrlDiffRmseHierNetNoise{})}\\
\textsc{glinternet} & \knobAucGlinternetInt{} {\scriptsize(\knobCtrlDiffAucGlinternetInt{})} & \knobFOneGlinternetInt{} {\scriptsize(\knobCtrlDiffFOneGlinternetInt{})} & \knobRmseGlinternetInt{} {\scriptsize(\knobCtrlDiffRmseGlinternetInt{})} & \knobAucGlinternetNoise{} {\scriptsize(\knobCtrlDiffAucGlinternetNoise{})} & \knobFOneGlinternetNoise{} {\scriptsize(\knobCtrlDiffFOneGlinternetNoise{})} & \knobRmseGlinternetNoise{} {\scriptsize(\knobCtrlDiffRmseGlinternetNoise{})}\\
SuSiE & \knobAucSusieInt{} {\scriptsize(\knobCtrlDiffAucSusieInt{})} & \knobFOneSusieInt{} {\scriptsize(\knobCtrlDiffFOneSusieInt{})} & \knobRmseSusieInt{} {\scriptsize(\knobCtrlDiffRmseSusieInt{})} & \knobAucSusieNoise{} {\scriptsize(\knobCtrlDiffAucSusieNoise{})} & \knobFOneSusieNoise{} {\scriptsize(\knobCtrlDiffFOneSusieNoise{})} & \knobRmseSusieNoise{} {\scriptsize(\knobCtrlDiffRmseSusieNoise{})}\\
lasso & \knobAucLassoInt{} {\scriptsize(\knobCtrlDiffAucLassoInt{})} & \knobFOneLassoInt{} {\scriptsize(\knobCtrlDiffFOneLassoInt{})} & \knobRmseLassoInt{} {\scriptsize(\knobCtrlDiffRmseLassoInt{})} & \knobAucLassoNoise{} {\scriptsize(\knobCtrlDiffAucLassoNoise{})} & \knobFOneLassoNoise{} {\scriptsize(\knobCtrlDiffFOneLassoNoise{})} & \knobRmseLassoNoise{} {\scriptsize(\knobCtrlDiffRmseLassoNoise{})}\\
stability selection & \knobAucStabilityRefitInt{} {\scriptsize(\knobCtrlDiffAucStabilityRefitInt{})} & \knobFOneStabilityRefitInt{} {\scriptsize(\knobCtrlDiffFOneStabilityRefitInt{})} & \knobRmseStabilityRefitInt{} {\scriptsize(\knobCtrlDiffRmseStabilityRefitInt{})} & \knobAucStabilityRefitNoise{} {\scriptsize(\knobCtrlDiffAucStabilityRefitNoise{})} & \knobFOneStabilityRefitNoise{} {\scriptsize(\knobCtrlDiffFOneStabilityRefitNoise{})} & \knobRmseStabilityRefitNoise{} {\scriptsize(\knobCtrlDiffRmseStabilityRefitNoise{})}\\
\bottomrule
\end{tabular*}
\end{table}

\subsection{Real outcomes}
\label{app:real-y}

The \realyN{} tables are the attitude survey and the Swiss fertility data shipped with R \citep{Rlang}, the prostate cancer and Los Angeles ozone data \citep{hastie2009elements}, the media-influence experiment of \citet{tal2010testing}, the body-fat data \citep{johnson1996fitting} and the diabetes data of \citet{efron2004least}. They have $n$ from \realyNMin{} to \realyNMax{} rows and $p$ from \realyPMin{} to \realyPMax{} predictors. Each has a continuous, documented outcome. The \realyNStress{} tables \realyStressTables{} are included as collinearity stress tests. The tables did not pass through the inclusion and cleaning rules of Appendix~\ref{app:predictor-matrices}, so no correlation or near-constant rule was applied to them. Every column other than the outcome is a predictor, with two exceptions. The train/test indicator of the prostate data is dropped. From the body-fat data we drop the case number, the body density from which the outcome is computed, and three columns computed from other columns (the Brozek estimate of body fat, adiposity and fat-free weight). The media-influence table is also one of the \rxPoolN{} predictor matrices, where its outcome column is one of the predictors. The predictors are standardized as in the benchmarks, and the outcome is centered and scaled to unit variance, because the prior is calibrated for outcomes of unit scale.

The reference on every table is the MCMC sampler of Appendix~\ref{app:mcmc}. It runs \realyChains{} chains of \realyStepsMin{} to \realyStepsMax{} steps, the step count rising with $p$. Every run met the convergence criteria, with a largest $\hat R$ of \realyRhatMax{} and a smallest effective sample size of \realyEssMin{}. Predictive quality is measured on a seeded split that holds out \realyQuerySharePct{} of the rows. It is the mean negative log predictive density on the held-out rows for \textsc{LinearPFN}, for the sampler and for OLS on the main effects, with the held-out rows standardized by the statistics of the fitting rows.

Pooled over the \realyMpmSlots{} candidate effects of the \realyN{} tables, the PIPs of \textsc{LinearPFN} correlate with those of the sampler at \realyPipRAll{}, at \realyPipRMains{} on the main effects and at \realyPipRInts{} on the interactions, with an RMSE of \realyPipRmseAll{}. The posterior-mean coefficients correlate at \realyCoefRHeadAll{} with an RMSE of \realyCoefRmseHeadAll{}. The two MPMs agree on \realyMpmSlotsSame{} of the \realyMpmSlots{} candidate effects. A Gaussian fitted to the fitting rows has a mean negative log predictive density of \realyNllMarginal{} on the held-out rows.

\begin{table}[t]
\caption{Real outcomes: \textsc{LinearPFN} against the MCMC sampler on \realyN{} published tables with their actual outcome. PIP agreement over the candidate effects (correlation on the main effects, RMSE and largest difference over all effects), RMSE between the posterior-mean coefficients, candidate effects on which the two MPMs agree, and held-out negative log predictive density.}
\label{tab:real-y}
\centering
\footnotesize
\setlength{\tabcolsep}{3pt}
\begin{tabular*}{\textwidth}{@{}l@{\extracolsep{\fill}}rrrrrrrrrr@{}}
\toprule
& & & \multicolumn{3}{c}{PIP} & coefficient & MPM & \multicolumn{3}{c}{held-out NLL}\\
\cmidrule(lr){4-6}\cmidrule(l){9-11}
table & $n$ & $p$ & $r$, mains & RMSE & max diff. & RMSE & agree & \textsc{LinearPFN} & MCMC & OLS\\
\midrule
attitude & \realyNAttitude{} & \realyPAttitude{} & \realyPipRMainsAttitude{} & \realyPipRmseAllAttitude{} & \realyPipMaxAbsAllAttitude{} & \realyCoefRmseHeadAttitude{} & \realyMpmSameAttitude{}/\realyMpmSlotsAttitude{} & \realyNllOursAttitude{} & \realyNllRefAttitude{} & \realyNllOlsAttitude{}\\
swiss & \realyNSwiss{} & \realyPSwiss{} & \realyPipRMainsSwiss{} & \realyPipRmseAllSwiss{} & \realyPipMaxAbsAllSwiss{} & \realyCoefRmseHeadSwiss{} & \realyMpmSameSwiss{}/\realyMpmSlotsSwiss{} & \realyNllOursSwiss{} & \realyNllRefSwiss{} & \realyNllOlsSwiss{}\\
prostate & \realyNProstate{} & \realyPProstate{} & \realyPipRMainsProstate{} & \realyPipRmseAllProstate{} & \realyPipMaxAbsAllProstate{} & \realyCoefRmseHeadProstate{} & \realyMpmSameProstate{}/\realyMpmSlotsProstate{} & \realyNllOursProstate{} & \realyNllRefProstate{} & \realyNllOlsProstate{}\\
tal\_or & \realyNTalOr{} & \realyPTalOr{} & \realyPipRMainsTalOr{} & \realyPipRmseAllTalOr{} & \realyPipMaxAbsAllTalOr{} & \realyCoefRmseHeadTalOr{} & \realyMpmSameTalOr{}/\realyMpmSlotsTalOr{} & \realyNllOursTalOr{} & \realyNllRefTalOr{} & \realyNllOlsTalOr{}\\
bodyfat & \realyNBodyfat{} & \realyPBodyfat{} & \realyPipRMainsBodyfat{} & \realyPipRmseAllBodyfat{} & \realyPipMaxAbsAllBodyfat{} & \realyCoefRmseHeadBodyfat{} & \realyMpmSameBodyfat{}/\realyMpmSlotsBodyfat{} & \realyNllOursBodyfat{} & \realyNllRefBodyfat{} & \realyNllOlsBodyfat{}\\
ozone & \realyNOzone{} & \realyPOzone{} & \realyPipRMainsOzone{} & \realyPipRmseAllOzone{} & \realyPipMaxAbsAllOzone{} & \realyCoefRmseHeadOzone{} & \realyMpmSameOzone{}/\realyMpmSlotsOzone{} & \realyNllOursOzone{} & \realyNllRefOzone{} & \realyNllOlsOzone{}\\
diabetes & \realyNDiabetes{} & \realyPDiabetes{} & \realyPipRMainsDiabetes{} & \realyPipRmseAllDiabetes{} & \realyPipMaxAbsAllDiabetes{} & \realyCoefRmseHeadDiabetes{} & \realyMpmSameDiabetes{}/\realyMpmSlotsDiabetes{} & \realyNllOursDiabetes{} & \realyNllRefDiabetes{} & \realyNllOlsDiabetes{}\\
\bottomrule
\end{tabular*}
\end{table}

\subsection{Training without the predictive loss}
\label{app:two-head}

The regression head is the only source of the posterior predictive distribution, and the headroom and coverage criteria score it (Appendix~\ref{app:validation}). A separate question is whether its loss also helps the other two heads. We retrained the paper's model with the predictive negative log-likelihood removed from Equation~\ref{eq:loss} and everything else unchanged, with the same prior, architecture, seed and number of steps (\trainSteps{}). Each of the two models was trained once. The selection and coefficient heads of this two-head model learn from the simulated labels alone. On the \ipDatasetsOursRealX{} datasets of the main benchmark, the paper's model minus the two-head model is \ipDiffAucLinearpfnStrongNonllRealX{} in AUC per dataset ($p \ipPAucLinearpfnStrongNonllRealX{}$), \ipDiffFOneLinearpfnStrongNonllRealX{} in $F_1$ per dataset ($p \ipPFOneLinearpfnStrongNonllRealX{}$) and \ipDiffRmseLinearpfnStrongNonllRealX{} in coefficient RMSE ($p \ipPRmseLinearpfnStrongNonllRealX{}$), with Wilcoxon signed-rank $p$ values. The paper's model has the smaller coefficient error on \ipWinRmseLinearpfnStrongNonllRealX{} of the datasets. The two-head model still has a higher $F_1$ than every baseline, \ipFOneMeanLinearpfnStrongNonllRealX{} against \ipFOneMeanSusieRealX{} for \textsc{SuSiE}. On the validation panel of Appendix~\ref{app:validation}, the two-head model meets the three selection criteria. Its AUC gap to the exact posterior is \gateAucGapArmVTwoMixFiveZeroNTwoKStrongNonllS{}, against \gateAucGap{} for the paper's model. Its ECE gap is \gateEceGapArmVTwoMixFiveZeroNTwoKStrongNonllS{}, against \gateEceGap{}. Its smallest resolution ratio, on the interactions, is \gateResRatioArmVTwoMixFiveZeroNTwoKStrongNonllS{}, against \gateResRatio{}. It cannot meet the headroom and coverage criteria, because its regression head is never trained. The validation does not score the coefficient head of the two-head model. The remaining two loss terms were not reweighted, so part of the gap may be recoverable.

\end{document}